\documentclass[runningheads]{llncs}

\usepackage{eccv}
\usepackage[width=122mm,left=14mm,paperwidth=150mm,height=193mm,top=12mm,paperheight=217mm]{geometry}

\usepackage{eccvabbrv}

\usepackage{graphicx}
\usepackage{booktabs}
\usepackage[ruled,linesnumbered]{algorithm2e}
\usepackage{algpseudocode}

\usepackage[accsupp]{axessibility}  

\usepackage{multirow}
\usepackage{bbding} 
\usepackage{pifont} 

\usepackage{hyperref}

\usepackage{orcidlink}

\hypersetup{
    colorlinks=true,
}

\begin{document}

%

\title{MVSGaussian: Fast Generalizable Gaussian Splatting Reconstruction from Multi-View Stereo
}


\titlerunning{MVSGaussian}

\author{
Tianqi Liu$^{1}$ \and
Guangcong Wang$^{2,3}$ \and
Shoukang Hu$^{2}$ \and
Liao Shen$^{1}$ \and \\
Xinyi Ye$^{1}$ \and
Yuhang Zang$^{4}$ \and
Zhiguo Cao$^{1\ast}$ \and
Wei Li$^{2\dag}$ \and
Ziwei Liu$^{2}$
}

\authorrunning{T.~Liu~et al.}

\institute{Huazhong University of Science and Technology \and
S-Lab, Nanyang Technological University \and
Great Bay University \and
Shanghai AI Laboratory \\
\email{\{tq\_liu,zgcao\}@hust.edu.cn} \\
\url{https://mvsgaussian.github.io/}
}

\maketitle
\let\thefootnote\relax\footnotetext{$^{\ast}$ Corresponding author}
\let\thefootnote\relax\footnotetext{$^{\dag}$ Project lead}

\begin{abstract}
We present MVSGaussian, a new generalizable 3D Gaussian representation approach derived from Multi-View Stereo (MVS) that can efficiently reconstruct unseen scenes.
Specifically, 1) we leverage MVS to encode geometry-aware Gaussian representations and decode them into Gaussian parameters. 2) To further enhance performance, we propose a hybrid Gaussian rendering that integrates an efficient volume rendering design for novel view synthesis. 3) To support fast fine-tuning for specific scenes, we introduce a multi-view geometric consistent aggregation strategy to effectively aggregate the point clouds generated by the generalizable model, serving as the initialization for per-scene optimization. Compared with previous generalizable NeRF-based methods, which typically require minutes of fine-tuning and seconds of rendering per image, MVSGaussian achieves real-time rendering with better synthesis quality for each scene. Compared with the vanilla 3D-GS, MVSGaussian achieves better view synthesis with less training computational cost. Extensive experiments on DTU, Real Forward-facing, NeRF Synthetic, and Tanks and Temples datasets validate that MVSGaussian attains state-of-the-art performance with convincing generalizability, real-time rendering speed, and fast per-scene optimization.

\end{abstract}
\begin{figure}[tb]
  \centering
    \begin{subfigure}{0.28\textwidth}
      \centering
      \includegraphics[width=1\linewidth]{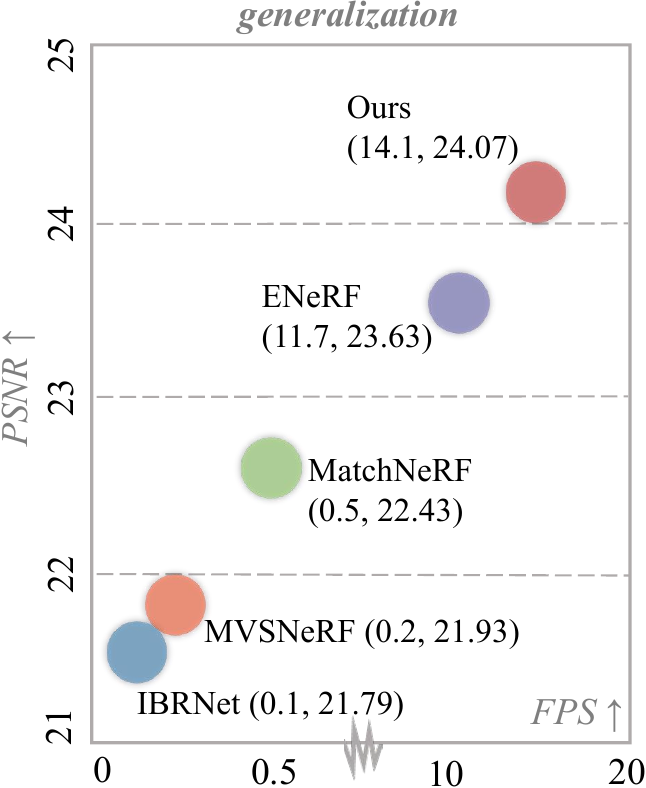}
      \caption{}
    \end{subfigure}
    \hfil
    \begin{subfigure}{0.328\textwidth}
      \centering
      \includegraphics[width=1\linewidth]{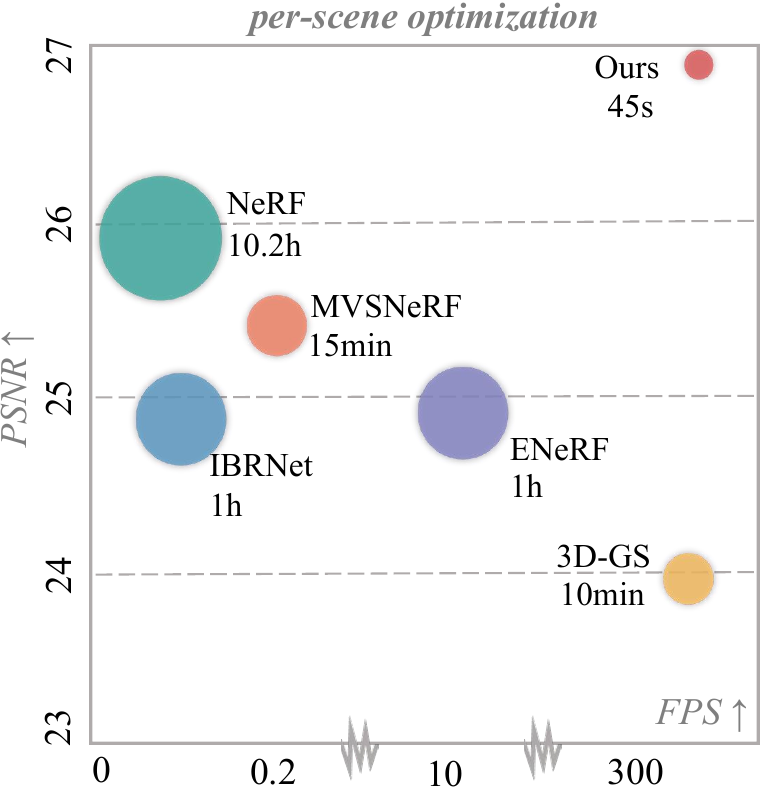}
      \caption{}
    \end{subfigure}
    \hfil
     \begin{subfigure}{0.333\textwidth}
      \centering
        \includegraphics[width=1\linewidth]{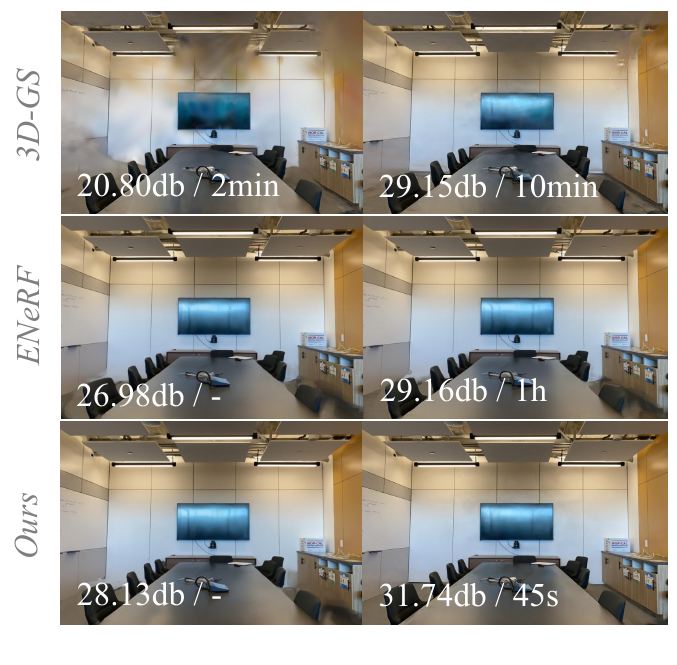}
          \caption{}
    \end{subfigure}
  \caption{\textbf{Comparison with existing methods.} (a) We present the generalizable results on the Real Forward-facing dataset~\cite{llff}. Compared with other competitors, our method achieves better performance at a faster inference speed. (b) The results after per-scene optimization, where circle size represents optimization time. Our method achieves optimal performance in just $45$ seconds. (c) We illustrate a scene (``room''), showcasing the (PSNR/optimization time) of synthesized views, with "-" indicating results from direct inference using the generalizable model.}
  \label{fig:fig1}
\end{figure}

\section{Introduction}
\label{sec:intro}
Novel view synthesis (NVS) aims to produce realistic images at novel viewpoints from a set of source images. By encoding scenes into implicit radiance fields, NeRF~\cite{nerf} has achieved remarkable success. However, this implicit representation is time-consuming due to the necessity of querying dense points for rendering.
Recently, 3D Gaussian Splatting (3D-GS)~\cite{3Dgaussians} utilizes anisotropic 3D Gaussians to explicitly represent scenes, achieving real-time and high-quality rendering through a differentiable tile-based rasterizer. However, 3D-GS relies on per-scene optimization for several minutes, which limits its applications. 


To remedy this issue, some initial attempts have been made to generalize Gaussian Splatting to unseen scenes. Generalizable Gaussian Splatting methods directly regress Gaussian parameters in a feed-forward manner instead of per-scene optimization. The general paradigm involves encoding features for 3D points in a scene-agnostic manner, followed by decoding these features to obtain Gaussian parameters. PixelSplat~\cite{pixelsplat} leverages an epipolar Transformer~\cite{Attention} to address scale ambiguity and encode features. However, it focuses on image pairs as input, and the introduction of Transformers results in significant computational overhead.
GPS-Gaussian~\cite{gpsgaussian} draws inspiration from stereo matching by first performing epipolar rectification on input image pairs, followed by disparity estimation and feature encoding. However, it focuses on human novel view synthesis and requires ground-truth depth maps. Splatter Image~\cite{splatterimage} introduces a single-view reconstruction approach based on Gaussian Splatting but focuses on object-centric reconstruction rather than generalizing to unseen scenes.

Due to the inefficiency of existing methods and their limitation to object-centric reconstruction, in this paper, we aim to develop an efficient generalizable Gaussian Splatting framework for novel view synthesis in unseen general scenes, which faces several critical challenges: 
\textbf{\textit{First}}, unlike NeRFs that use an implicit representation, 3D-GS is a parameterized explicit representation that uses millions of 3D Gaussians to overfit a scene. When applying the pre-trained 3D-GS to an unseen scene, the parameters of 3D Gaussians, such as locations and colors, are significantly different. It is a non-trivial problem to design a generalizable representation to tailor 3D-GS.
\textbf{\textit{Second}}, previous generalizable NeRFs~\cite{pixelnerf,ibrnet,mvsnerf,enerf,matchnerf} have achieved impressive view synthesis results through volume rendering. However, the generalization capability of splatting remains unexplored. During splatting, each Gaussian contributes to multiple pixels within a certain region in the image, and each pixel's color is determined by the accumulated contributions from multiple Gaussians. The color correspondence between Gaussians and pixels is a more complex many-to-many mapping, which poses a challenge for model generalization.
\textbf{\textit{Third}}, generalizable NeRFs show that further fine-tuning for specific scenes can greatly improve the synthesized image quality but requires lengthy optimization. Although 3D-GS is faster than NeRF, it still remains time-consuming. Designing a fast optimization approach based on the generalizable 3D-GS model is promising.

We address these challenges point by point. \textbf{\textit{First}}, we propose leveraging MVS for geometry reasoning and encoding features for 3D points to establish pixel-aligned Gaussian representations. The point-wise features are aggregated from multi-view features, and the spatial awareness is enhanced through a 2D UNet, as each Gaussian contributes to multiple pixels. \textbf{\textit{Second}}, with the encoded point-wise features, we can decode them into Gaussian parameters through an MLP. Rather than solely relying on splatting, we propose adding a simple yet effective depth-aware volume rendering approach to enhance generalization. \textbf{\textit{Third}}, with the trained generalizable model, lots of 3D Gaussians can be generated from multiple views. These Gaussian point clouds can serve as an initialization for subsequent per-scene optimization. However, the generated 3D Gaussians from the generalizable model are not perfect. Directly concatenating such a large number of Gaussians as initialization for per-scene optimization leads to unexpected computational costs because these Gaussians further split and clone during optimization.
One approach is to downsample the point cloud, such as voxel downsampling, which can reduce noise but also result in the loss of effective information.
Therefore, we introduce a strategy to aggregate point clouds by preserving multi-view geometric consistency. Specifically, we filter out noisy points by computing the reprojection error of the depth of Gaussians from different viewpoints. This strategy can filter out noisy points while preserving effective ones, providing a high-quality initialization for subsequent optimization.

To summarize, we present a new fast generalizable Gaussian Splatting method. 
We evaluate our method on the widely-used DTU~\cite{dtu}, Real Forward-facing~\cite{llff}, NeRF Synthetic~\cite{nerf}, and Tanks and Temples~\cite{tanks} datasets. Extensive experiments show that our generalizable method outperforms other generalizable methods. After a short period of per-scene optimization, our method achieves performance comparable to or even better than other methods with longer optimization times, as shown in Fig.~\ref{fig:fig1}. 
On a single RTX 3090 GPU, compared with the vanilla 3D-GS, our proposed method achieves better novel view synthesis with similar rendering speed ($300$+ FPS) and $13.3$× less training computational cost ($45$s). 
\noindent Our main contributions can be summarized as follows: 
\begin{itemize}
    \item We present MVSGaussian, a generalizable Gaussian Splatting method derived from Multi-View Stereo and a pixel-aligned Gaussian representation.
    \item We further propose an efficient hybrid Gaussian rendering approach to boost generalization learning.
    \item We introduce a consistent aggregation strategy to provide high-quality initialization for fast per-scene optimization.
\end{itemize}

\section{Related Work}
\label{sec:related work}
\noindent \textbf{Multi-View Stereo (MVS)} aims to reconstruct a dense 3D representation from multiple views. Traditional MVS methods~\cite{fua1995object,gipuma,colmap,schonberger2016pixelwise} rely on hand-crafted features and similarity metrics, which limits their performance. With the advancement of deep learning in 3D perception, MVSNet~\cite{mvsnet} first proposes an end-to-end pipeline, with the key idea being the construction of a cost volume to aggregate 2D information into a 3D geometry-aware representation. Subsequent works follow this cost volume-based pipeline and make improvements from various aspects, \eg reducing memory consumption with recurrent plane sweeping~\cite{D2HC-RMVSNet,rmvsnet}  or coarse-to-fine architectures~\cite{cheng2020deep,gu2020cascade,yang2020cost}, optimizing cost aggregation~\cite{aa-rmvsnet,mvster}, enhancing feature representations~\cite{transmvsnet,et-mvsnet}, and improving decoding strategy~\cite{unimvs,dmvsnet}. As the cost volume encodes the consistency of multi-view features and naturally performs correspondence matching, in this paper, we develop a new generalizable Gaussian Spatting representation derived from MVS.

\noindent \textbf{Generalizable NeRF.}
By implicitly representing scenes as continuous color and density fields using MLPs, Neural Radiance Fields (NeRF) achieve impressive rendering results with volume rendering techniques. Follow-up works~\cite{sofgan,neutex,nerd,nerfies,xian2021space,sparsenerf,irshad2023neo360,perf2023,consistentnerf} extend it to various tasks and achieve promising results. However, they all require time-consuming per-scene optimization. To address this issue, some generalizable NeRFs have been proposed. The general paradigm involves encoding features for each 3D point and then decoding these features to obtain volume density and radiance. According to the encoded features, generalizable NeRFs can be categorized into appearance features~\cite{pixelnerf}, aggregated multi-view features~\cite{ibrnet,enerf,liu2024gefu,gnt}, cost volume-based features~\cite{mvsnerf,neuray,enerf,liu2024gefu}, and correspondence matching features~\cite{matchnerf}. Despite considerable progress, performance remains limited, with slow optimization and rendering speeds.

\noindent \textbf{3D Gaussian Splatting (3D-GS)} utilizes anisotropic Gaussians to explicitly represent scenes and achieves real-time rendering through differentiable rasterization. Motivated by this, several studies have applied it to various tasks, \eg editing~\cite{Gaussianeditor,cen2023saga}, dynamic scenes~\cite{yang2023deformable3dgs,luiten2023dynamic,wu20234dgaussians}, avatars~\cite{GauHuman,GaussianAvatar,3DGS-Avatar} and others~\cite{chen2024survey}. However, the essence of Gaussian Splatting still lies in overfitting the scene. To remedy this, a few concurrent works make initial attempts to generalize Gaussian Splatting to unseen scenes. The goal of Generalizable Gaussian Splatting is to predict Gaussian parameters in a feed-forward manner instead of per-scene optimization. PixelSplat~\cite{pixelsplat} addresses scale ambiguity by leveraging an epipolar Transformer to encode features and subsequently decode them into Gaussian parameters. However, it focuses on image pairs as input and the Transformer incurs significant computational costs. GPS-Gaussian~\cite{gpsgaussian} draws inspiration from stereo matching and performs epipolar rectification and disparity estimation on input image pairs. However, it focuses on human novel view synthesis and requires ground-truth depth maps. Spatter Image~\cite{splatterimage} introduces a single-view 3D reconstruction approach. However, it focuses on object-centric reconstruction rather than generalizing to unseen scenes. Overall, these methods are constrained by inefficiency, limited to object reconstruction, and restricted to either image pairs or a single view. To this end, in this paper, we aim to study an efficient generalizable Gaussian Splatting for novel view synthesis in unseen general scenes.

\section{Preliminary}
\label{sec:preliminary}
\textbf{3D Gaussian Splatting} represents a 3D scene as a mixture of anisotropic 3D Gaussians, each of which is defined with a 3D covariance matrix $\Sigma$ and mean $\mu$:
\begin{equation}
\label{eq:1}
G(X) = e^{-\frac{1}{2}(X-\mu)^T \Sigma^{-1} (X-\mu)}\,.
\end{equation}
The covariance matrix $\Sigma$ holds physical meaning only when it is positive semi-definite. Therefore, for effective optimization through gradient descent, $\Sigma$ is decomposed into a scaling matrix $S$ and a rotation matrix $R$, as $\Sigma = RSS^TR^T$.
To splat Gaussians from 3D space to a 2D plane, the view transformation $W$ and the Jacobian matrix $J$ representing the affine approximation of the projective transformation are utilized to obtain the covariance matrix $\Sigma'$ in 2D space, as $\Sigma' = JW\Sigma W^T J^T$.
Subsequently, a point-based alpha-blend rendering can be performed to obtain the color of each pixel:
\begin{equation}
\label{eq:4}
C = \sum_{i} c_i \alpha_i{\prod_{j=1}^{i-1} (1 - \alpha_i)}\,,
\end{equation}
where $c_i$ is the color of each point, defined by spherical harmonics (SH) coefficients. The density $\alpha_i$ is computed by the multiplication of 2D Gaussians and a learnable point-wise opacity. During optimization, the learnable attributes of each Gaussian are updated through gradient descent, including 1) a 3D position $\mu \in \mathbb{R}^3$, 2) a scaling vector $s \in \mathbb{R}_{+}^3$, 3) a quaternion rotation vector $r \in \mathbb{R}^4$, 4) a color defined by SH $c \in \mathbb{R}^k$ (where k is the freedom), 
and 5) an opacity $\alpha \in [0, 1]$. Additionally, an adaptive density control module is introduced to improve rendering quality, comprising mainly the following three operations: 1) split into
smaller Gaussians if the magnitude of the scaling exceeds a threshold, 2) clone if the magnitude of the scaling is smaller than a threshold, and 3) prune Gaussians with excessively small opacity or overly large scaling magnitudes.
\section{MVSGaussian}
\label{sec:method}

\begin{figure}[tb]
  \centering
  \includegraphics[width=1\textwidth]{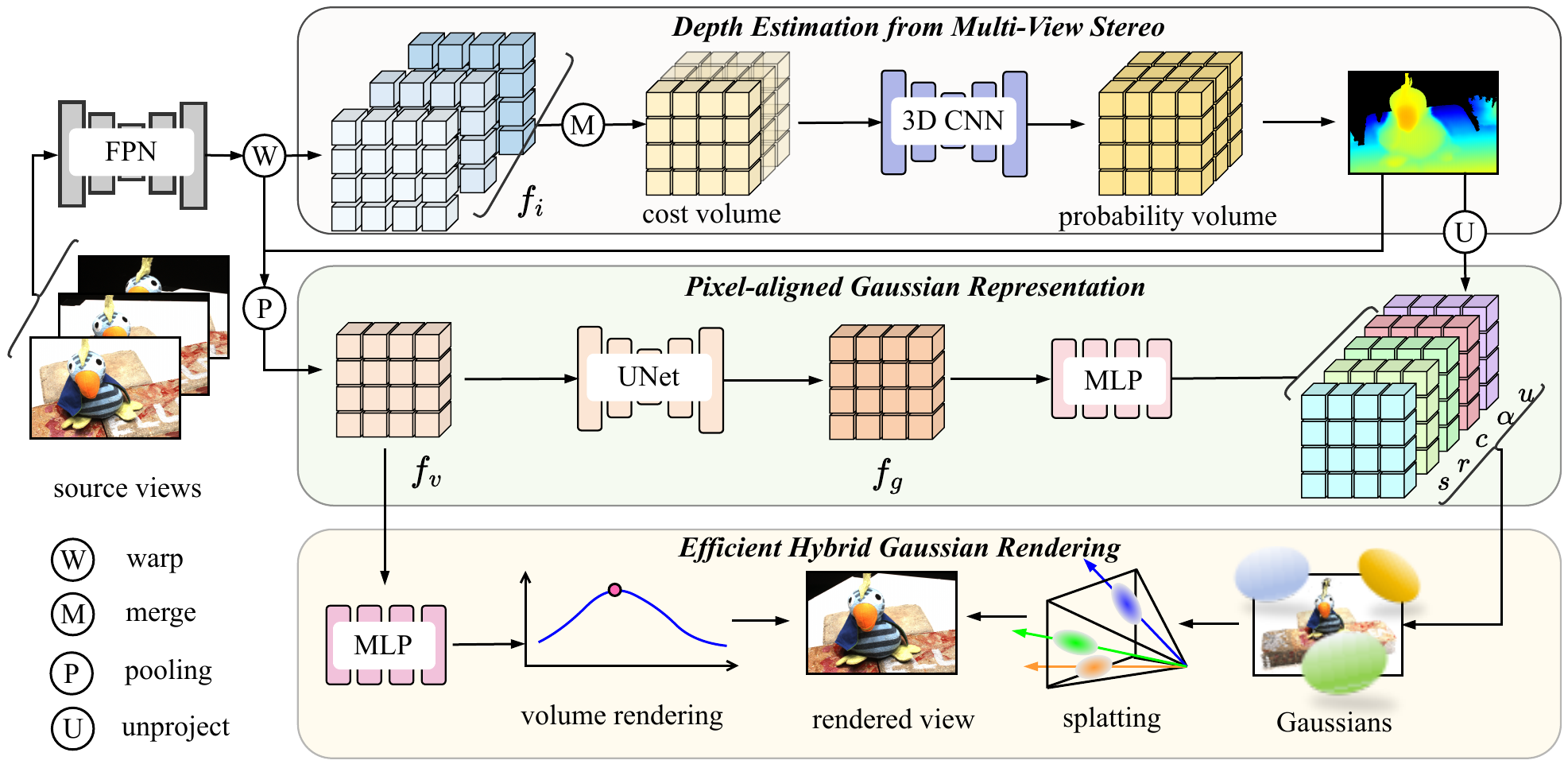}
  \caption{\textbf{Overview of MVSGaussian.} We first extract features $\{f_i\}_{i=1}^N$ from input source views $\{I_i\}_{i=1}^N$ using FPN. These features are then aggregated into a cost volume, regularized by 3D CNNs to produce depth. Subsequently, for each 3D point at the estimated depth, we use a pooling network to aggregate warped source features, obtaining the aggregated feature $f_v$. This feature is then enhanced using a 2D UNet, yielding the enhanced feature $f_g$. $f_g$ is decoded into Gaussian parameters for splatting, while $f_v$ is decoded into volume density and radiance for depth-aware volume rendering. Finally, the two rendered images are averaged to produce the final rendered result.}
  \label{fig:pipeline}
\end{figure}

\subsection{Overview} 

Given a set of source views $\{I_i\}_{i=1}^N$, NVS aims to synthesize a target view from a novel camera pose. The overview of our proposed generalizable Gaussian Splatting framework is depicted in Fig.~\ref{fig:pipeline}. We first utilize a Feature Pyramid Network (FPN)~\cite{fpn} to extract multi-scale features from source views. These features are then warped onto the target camera frustum to construct a cost volume via differentiable homography, followed by 3D CNNs for regularization to produce the depth map. Based on the obtained depth map, we encode features for each pixel-aligned 3D point by aggregating multi-view and spatial information. The encoded features can be then decoded for rendering. However, Gaussian Splatting is a region-based explicit representation and is designed for tile-based rendering, involves a complex many-to-many mapping between Gaussians and pixels, posing challenges for generalizable learning. To address this, we propose an efficient hybrid rendering by integrating a simple depth-aware volume rendering module, where only one point is sampled per ray. We render two views using Gaussian Splatting and volume rendering, then average these two rendered views into the final view. This pipeline is further constructed in a cascade structure, propagating the depth map and rendered view in a coarse-to-fine manner.

\subsection{MVS-based Gaussian Splatting Representation}
\label{subsec:mvs-rep}

\textbf{Depth Estimation from MVS.} The depth map is a crucial component of our pipeline, as it bridges 2D images and 3D scene representation. Following learning-based MVS methods~\cite{mvsnet}, we first establish multiple fronto-parallel planes at the target view. Then, we warp the features of source views onto these sweeping planes using differentiable homography as:
\begin{equation}
\label{eq:homography}
H_i(z) = K_i R_i (I + \frac{(R_i^{-1} t_i - R_t^{-1} t_t) a^T R_t}{z}) R_t^{-1} K_t^{-1}\,,
\end{equation}
where $[K_i,R_i,t_i]$ and $[K_t,R_t,t_t]$ are the camera intrinsic, rotation and translation of the source view $I_i$ and target view, respectively. The $a$ represents the principal axis of the target view camera, $I$ denotes the identity matrix and $z$ is the sampled depth. With the warped features from source views, a cost volume is constructed by computing their variance, which encodes the consistency of multi-view features. Then, the cost volume is fed into 3D CNNs for regularization to obtain the probability volume. With this depth probability distribution, we weight each depth hypothesis to obtain the final depth.

\noindent\textbf{Pixel-aligned Gaussian Representation.} 
With the estimated depth, each pixel can be unprojected to a 3D point, which is the position of the 3D Gaussian. The subsequent step is encoding features for these 3D points to establish a pixel-aligned Gaussian representation. Specifically, we first warp the features from source views to the target camera frustum using~\cref{eq:homography}, and then utilize a pooling network $\rho$~\cite{ibrnet,enerf} to aggregate these multi-view features into features $f_v = \rho(\{f_i\}_{i=1}^N)$. Considering the properties of splatting, each Gaussian contributes to the color values of pixels in a specific region of the image. However, the aggregated feature $f_v$ only encodes multi-view information for individual pixels, lacking spatial awareness. Therefore, we utilize a 2D UNet for spatial enhancement, yielding $f_g$.
With the encoded features, we can decode them to obtain Gaussian parameters for rendering. Specifically, each Gaussian is characterized by attributes $\{\mu,s,r,\alpha,c\}$ as described in Sec. \ref{sec:preliminary}.
For the position $\mu$, it can be obtained by unprojecting pixels according to the estimated depth as:
\begin{equation}
\label{eq:mu}
\mu = \Pi^{-1} (x,d)\,,
\end{equation}
where $\Pi^{-1}$ represents the unprojection operation. $x$ and $d$ represent the coordinates and estimated depth of the pixel, respectively. For scaling $s$, rotation $r$, and opacity $\alpha$, they can be decoded from the encoded features, given by:
\begin{equation}
\label{eq:scale}
\begin{aligned}
s &= Softplus(h_s(f_g))\,, \\
r &= Norm(h_r(f_g))\,, \\
\alpha &= Sigmoid(h_{\alpha}(f_g))\,,
\end{aligned}
\end{equation}
where $h_s$, $h_r$, and $h_\alpha$ represent the scaling head, rotation head, and opacity head, respectively, instantiated as MLPs.
For the last attribute, color $c$, 3D Gaussian Splatting~\cite{3Dgaussians} utilizes spherical harmonic (SH) coefficients to define it. However, the generalization of learning SH coefficients from features is not robust~(\cref{subsec:ablations and analysis}). Instead, we directly regress color from features as:
\begin{equation}
\label{eq:color}
c = Sigmoid(h_c (f_g))\,,
\end{equation}
where $h_c$ represents the color head.

\noindent\textbf{Efficient Hybrid Gaussian Rendering.} With the aforementioned Gaussian parameters, a novel view can be rendered using the splatting technique. However, the obtained view lacks fine details, and this approach exhibits limited generalization performance. Our insight is that the splatting approach introduces a complex many-to-many relationship between 3D Gaussians and pixels in terms of color contribution, which poses challenges for generalization. Therefore, we propose using a simple one-to-one correspondence between 3D Gaussians and pixels to predict colors for refinements. In this case, the splatting degenerates into the volume rendering with a single depth-aware sampling point. Specifically, following~\cite{enerf,ibrnet}, we obtain radiance and volume density by decoding $f_v$, followed by volume rendering to obtain a rendered view. The final rendered view is formed by averaging the views rendered through splatting and volume rendering.

\subsection{Consistent Aggregation for Per-Scene Optimization}
\label{subsec: per-scene optimization}

\begin{figure}[tb]
  \centering
  \includegraphics[width=1\textwidth]{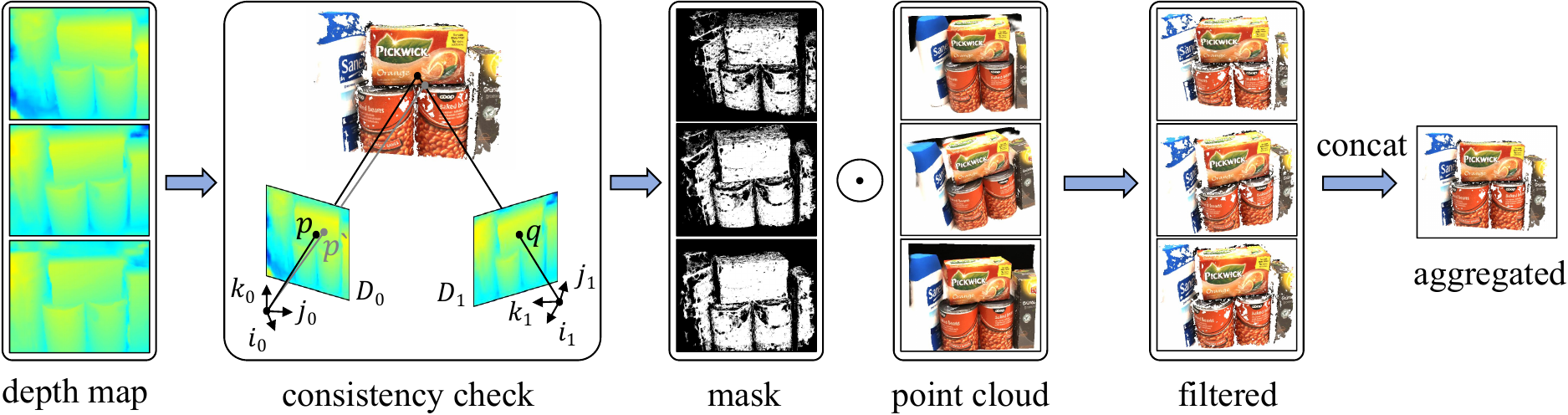}
  \caption{\textbf{Consistent aggregation.} With depth maps and point clouds produced by the generalizable model, we first conduct geometric consistency checks on depths to derive masks for filtering out unreliable points. The filtered point clouds are then concatenated to obtain a point cloud, serving as the initialization for per-scene optimization.}
  \label{fig:fusion}
\end{figure}

The generalizable model can reconstruct a reasonable 3D Gaussian representation for an unseen scene. We can further optimize this Gaussian representation for specific scenes using optimization strategies described in~\cref{sec:preliminary}. Since the aforementioned generalizable model reconstructs Gaussian representations at several given novel viewpoints, the primary challenge is how to effectively aggregate these Gaussian representations into a single Gaussian representation for efficient rendering. Due to the inherent limitations of the MVS method, the depth predicted by the generalizable model may not be entirely accurate, leading to the presence of noise in the resulting Gaussian point cloud. Directly concatenating these Gaussian point clouds results in a significant amount of noise. Additionally, a large number of points slow down subsequent optimization and rendering speeds. An intuitive solution is to downsample the concatenated point cloud. However, while reducing noise, it also diminishes the number of effective points. Our insight is that a good aggregation strategy should minimize noisy points and retain effective ones as much as possible, while also ensuring that the total number of points is not excessively large. To this end, we introduce an aggregation strategy based on multi-view geometric consistency. The predicted depth for the same 3D point across different viewpoints should demonstrate consistency. Otherwise, the predicted depth is considered unreliable. This geometric consistency can be measured by calculating the reprojection error between different views. Specifically, as illustrated in Fig~\ref{fig:fusion}, given a reference depth map $D_0$ to be examined and a depth map $D_1$ from a nearby viewpoint, we first project the pixel $p$ in $D_0$ to the nearby view to obtain the projected point $q$ as:
\begin{equation}
\label{eq:0-1}
q = \frac{1}{d} \Pi_{0-1} (p,D_0(p))\,,
\end{equation}
where $\Pi_{0-1}$ represents the transformation from $D_0$ to $D_1$, and $d$ is the depth from projection. In turn, we back-project the obtained pixel $q$ with estimated depth
$D_1(q)$ onto the reference view to obtain the reprojected point $p'$ as:
\begin{equation}
\label{eq:1-0}
p' = \frac{1}{d'} \Pi_{1-0} (q,D_1(q))\,,
\end{equation}
where $\Pi_{1-0}$ represents the transformation from $D_1$ to $D_0$, and $d'$ is the depth of the reprojected pixel. Then, the reprojection errors are calculated by:
\begin{equation}
\label{eq:re}
\begin{aligned}
\xi_p &= \left\|{p - p'}\right\|_2 \,, \\
\xi_d &= {\left\|{D_0(p) - d'}\right\|_1}/{D_0(p)}\,,
\end{aligned}
\end{equation}
The reference image will be compared pairwise with each of the remaining images to calculate the reprojection error. Inspired by~\cite{yan2020dense,et-mvsnet}, we adopt the dynamic consistency checking algorithm to select the valid depth
values. The main idea is that the estimated depth is reliable when it has very a low reprojection error in a minority of views or a relatively low error in the majority of views. It can be formulated as follows:
\begin{equation}
\label{eq:dypcd}
\xi_p < \theta_p(n)\,, \xi_d < \theta_d(n) \,,
\end{equation}
where $\theta_p(n)$ and $\theta_d(n)$ represent predefined thresholds, whose values increase as the number of views $n$ increases. The depth is reliable when there are $n$ nearby views that meet the corresponding thresholds $\theta_p(n)$ and $\theta_d(n)$. We filter out noise points that do not meet the conditions and store the correctly reliable points.

\subsection{Full Objective}
\label{subsec: loss function}
Our model is trained end-to-end using only RGB images as supervision. We optimize the generalizable model with the mean squared error (mse) loss, SSIM loss~\cite{ssim}, and perceptual loss~\cite{lpips}, as follows:
\begin{equation}
\label{eq:k_th_loss}
L^{k}=L_{\mathrm{mse}} +\lambda_s L_{\mathrm{ssim}} +\lambda_p L_{\mathrm{perc}} \,,
\end{equation}
where $L^{k}$ represents the loss for the $k^{th}$ stage of the coarse-to-fine framework. $\lambda_s$ and $\lambda_p$ denote the loss weights.
The overall loss is the sum of losses from each stage, given by:
\begin{equation}
\label{eq:overall_loss}
L = \sum \lambda ^k  L^k \,,
\end{equation}
where $\lambda ^k$ represents the loss weight for the $k^{th}$ stage.
During per-scene optimization, following~\cite{3Dgaussians}, 
we optimize Gaussian point clouds using the $L_1$ loss combined with a D-SSIM term:
\begin{equation}
\label{eq:finetuning_loss}
L_{\mathrm{ft}} = (1-\lambda_{\mathrm{ft}})L_1 + \lambda_{\mathrm{ft}}L_{\mathrm{D-SSIM}} \,,
\end{equation}
where $\lambda_{\mathrm{ft}}$ is the loss weight.
\section{Experiments}
\label{sec:experiments}

\begin{table*}[t]
\caption{\textbf{Quantitative results of generalization on the DTU test set~\cite{dtu}.} FPS and Mem are measured under a 3-view input, while FPS$^*$ and Mem$^*$ are measured under a 2-view input. The best result is in \textbf{bold}, and the second-best one is in \underline{underlined}.}
\label{Tab:gen_dtu}
\centering
\setlength{\tabcolsep}{5pt}
\resizebox{1\linewidth}{!}{
\begin{tabular}{@{}lccccccccc@{}}
\toprule
\multirow{2}{*}{Method} & & \multicolumn{3}{c}{3-view} & \multicolumn{3}{c}{2-view} & \multirow{2}{*}{Mem (GB)$\downarrow$}  & \multirow{2}{*}{FPS$\uparrow$}\\ 
\cmidrule(lr){3-5} \cmidrule(lr){6-8}
 & & PSNR $\uparrow$ & SSIM $\uparrow$ & LPIPS $\downarrow$ & PSNR $\uparrow$ & SSIM $\uparrow$ & LPIPS $\downarrow$ \\
\midrule
PixelNeRF~\cite{pixelnerf} & & 19.31 & 0.789 & 0.382 & - & - & - & - & 0.019 \\
IBRNet~\cite{ibrnet} & & 26.04 & 0.917 & 0.191 & - & - & -  & - & 0.217 \\ 
MVSNeRF~\cite{mvsnerf}  & & 26.63 & 0.931 & 0.168 & 24.03 & 0.914 & 0.192 & - & 0.416 \\
ENeRF~\cite{enerf}  & & \underline{27.61} & \underline{0.957} & \underline{0.089} & \underline{25.48} & \underline{0.942} & \underline{0.107} & 2.183 & 19.5 \\
MatchNeRF~\cite{matchnerf} & & 26.91 & 0.934 & 0.159 & 25.03 & 0.919 & 0.181 & - & 1.04\\
PixelSplat~\cite{pixelsplat} & & - & - & - & 14.01 & 0.662 & 0.389 & 11.827$^{*}$ & 1.13$^{*}$ \\
Ours & & \textbf{28.21} & \textbf{0.963} & \textbf{0.076} & \textbf{25.78} & \textbf{0.947} & \textbf{0.095} & 0.876/0.866$^{*}$ & 21.5/24.5$^{*}$ \\
\bottomrule
\end{tabular}}
\end{table*}

\subsection{Settings}
\label{subsec:settings}
\noindent \textbf{Datasets.}
Following MVSNeRF~\cite{mvsnerf}, we train the generalizable model on the DTU training set~\cite{dtu} and evaluate it on the DTU test set. Subsequently, we conduct further evaluations on the Real Forward-facing~\cite{llff}, NeRF Synthetic~\cite{nerf}, and Tanks and Temples~\cite{tanks} datasets. For each test scene, we select $20$ nearby views, with 16 views comprising the working set and the remaining 4 views as testing views. The quality of synthesized views is measured by widely-used PSNR, SSIM~\cite{ssim}, and LPIPS~\cite{lpips} metrics.

\noindent \textbf{Baselines.}
We compare our method with state-of-the-art generalizable NeRF methods~\cite{pixelnerf,ibrnet,mvsnerf,enerf,matchnerf}, as well as the recent generalizable Gaussian method~\cite{pixelsplat}.
For the generalization comparison, we follow the same experimental settings as~\cite{mvsnerf,enerf,matchnerf} and borrow some results reported in~\cite{mvsnerf,matchnerf}. For~\cite{enerf} and~\cite{pixelsplat}, we evaluate them using their officially released code and pre-trained models. 
For per-scene optimization experiments, we include NeRF~\cite{nerf} and 3D-GS~\cite{3Dgaussians} for comparison.

\noindent \textbf{Implementation Details.}
Following~\cite{enerf}, we employ a two-stage cascaded framework. For depth estimation, we sample $64$ and $8$ depth planes for the coarse and fine stages, respectively.
We set $\lambda_s=0.1$ and $\lambda_p=0.05$ in \cref{eq:k_th_loss}, $\lambda^1=0.5$ and $\lambda^2=1$ in \cref{eq:overall_loss}, and $\lambda_{ft}=0.2$ in \cref{eq:finetuning_loss}. The generalizable model is trained using the Adam optimizer~\cite{adam} on four RTX 3090 GPUs. During the per-scene optimization stage, for fair comparison, our optimization strategy and hyperparameters settings remain consistent with the vanilla 3D-GS~\cite{3Dgaussians}, except for the number of iterations. For the initialization of 3D-GS, we use COLMAP~\cite{colmap} to reconstruct the point cloud from the working set.

\subsection{Generalization Results}
\label{subsec:generalization results}
\begin{figure}[tb]
  \centering
  \includegraphics[width=1\textwidth]{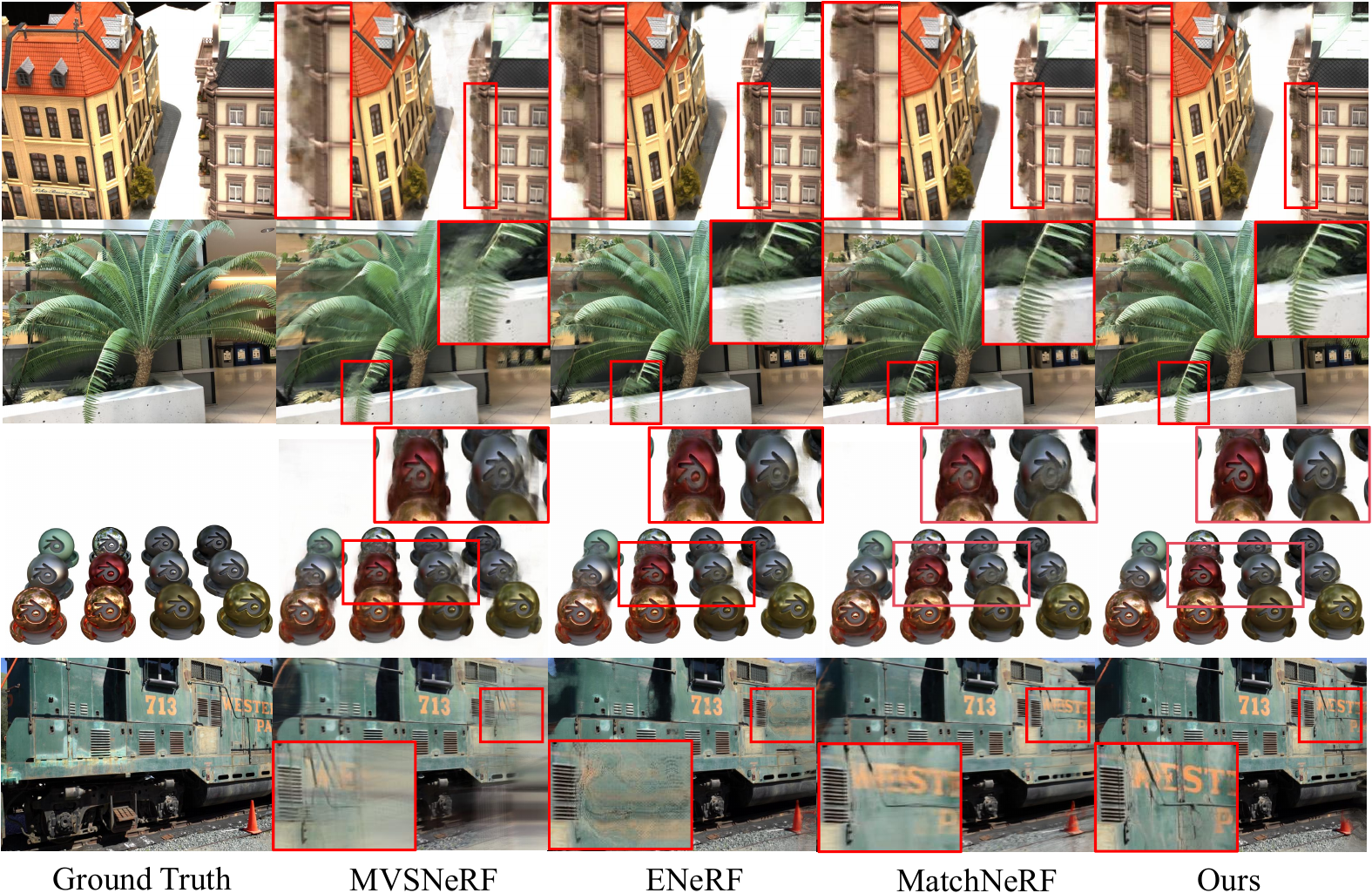}
  \caption{\textbf{Qualitative comparison of rendering quality under generalization and 3-view settings with state-of-the-art methods~\cite{mvsnerf,enerf,matchnerf}.} }
  \label{fig:vis_gen}
\end{figure}

We train the generalizable model on the DTU training set and report quantitative results on the DTU test set in Table~\ref{Tab:gen_dtu}, and the quantitative results on three additional datasets in Table~\ref{Tab:gen_llff}. Due to the MVS-based pixel-aligned Gaussian representation and the efficient hybrid Gaussian rendering, our method achieves optimal performance at a fast inference speed. Due to the introduction of the epipolar Transformer, PixelSplat~\cite{pixelsplat} has slow speed and large memory consumption. Additionally, it focuses on natural scenes with image pairs as input, and its performance significantly decreases when applied to object-centric datasets~\cite{dtu,nerf}. For NeRF-based methods, ENeRF~\cite{enerf} enjoys promising speeds by sampling only $2$ points per ray, however, its performance is limited and consumes higher memory overhead. The remaining methods render images by sampling rays due to their high memory consumption, as they cannot process the entire image at once. The qualitative results are presented in Fig.~\ref{fig:vis_gen}. Our method produces high-quality views with more scene details and fewer artifacts.

\begin{table}[t]
\caption{\textbf{Quantitative results of generalization on Real Forward-facing~\cite{llff}, NeRF Synthetic~\cite{nerf}, and Tanks and Temples~\cite{tanks} datasets.} Due to the significant memory consumption of PixelSplat~\cite{pixelsplat}, we conduct performance evaluation and comparison on low-resolution ($512\times512$) images, denoted as PixelSplat$^*$ and Ours$^*$. The best result is in \textbf{bold}, and the second-best one is in \underline{underlined}.}
\label{Tab:gen_llff}
\centering
\setlength{\tabcolsep}{5pt}
\resizebox{1\linewidth}{!}{
\begin{tabular}{@{}lccccccccccc@{}}
\toprule
\multirow{2}{*}{Method} & \multirow{2}{*}{Settings} & \multicolumn{3}{c}{Real Forward-facing~\cite{llff}} & \multicolumn{3}{c}{NeRF Synthetic~\cite{nerf}} & \multicolumn{3}{c}{Tanks and Temples~\cite{tanks}}\\ 
\cmidrule(lr){3-5}\cmidrule(lr){6-8}\cmidrule(lr){9-11}
 & & PSNR $\uparrow$ & SSIM $\uparrow$ & LPIPS $\downarrow$ & PSNR $\uparrow$ & SSIM $\uparrow$ & LPIPS $\downarrow$ & PSNR $\uparrow$ & SSIM $\uparrow$ & LPIPS $\downarrow$ \\
\midrule
PixelNeRF~\cite{pixelnerf} &\multirow{6}{*}{3-view} & 11.24 & 0.486 & 0.671 & 7.39 & 0.658 & 0.411 & - & - & - \\
IBRNet~\cite{ibrnet} & & 21.79 & 0.786 & 0.279 & 22.44 & 0.874 & 0.195 & 20.74 & 0.759 & 0.283 \\ 
MVSNeRF~\cite{mvsnerf}  & & 21.93 & 0.795 & 0.252 & 23.62 & 0.897 & 0.176 & 20.87 & 0.823 & 0.260 \\
ENeRF~\cite{enerf}  & & \underline{23.63} & \underline{0.843} & \underline{0.182} & \underline{26.17} & \underline{0.943} & \underline{0.085} & \underline{22.53} & \underline{0.854} & \underline{0.184} \\
MatchNeRF~\cite{matchnerf} & & 22.43 & 0.805 & 0.244 & 23.20 & 0.897 & 0.164 & 20.80 & 0.793 & 0.300 \\
Ours & & \textbf{24.07} & \textbf{0.857} & \textbf{0.164} & \textbf{26.46} & \textbf{0.948} & \textbf{0.071} & \textbf{23.28} & \textbf{0.877} & \textbf{0.139}  \\
\midrule
MVSNeRF~\cite{mvsnerf} & \multirow{4}{*}{2-view} & 20.22 & 0.763 & 0.287 & 20.56 & 0.856 & 0.243 & 18.92 & 0.756 & 0.326 \\
ENeRF~\cite{enerf} & &  \underline{22.78} & \underline{0.821} & \underline{0.191} & \underline{24.83} & \underline{0.931} & \underline{0.117} & \underline{22.51} & \underline{0.835} & \underline{0.193} \\
MatchNeRF~\cite{matchnerf} & & 20.59 & 0.775 & 0.276 & 20.57 & 0.864 & 0.200 & 19.88 & 0.773 & 0.334\\
Ours & & \textbf{23.11} & \textbf{0.834} & \textbf{0.175} & \textbf{25.06} & \textbf{0.937} & \textbf{0.079} & \textbf{22.67} & \textbf{0.844} & \textbf{0.162} \\
\midrule
PixelSplat$^{*}$~\cite{pixelsplat} & \multirow{2}{*}{2-view} & 22.99 & 0.810 & 0.190 & 15.77 & 0.755 & 0.314 & 19.40 & 0.689 & 0.223 \\
Ours$^{*}$ & & 23.30 & 0.835 & 0.152 & 25.34 & 0.935 & 0.071 & 23.18 & 0.849 & 0.130 \\
\bottomrule
\end{tabular}}
\end{table}
\begin{table}[t]
\caption{\textbf{Quantitative results after per-scene optimization.} Time$_{ft}$ represents the time for fine-tuning. The best result is in \textbf{bold}, and second-best one is in \underline{underlined}.}
\label{Tab:fine-tuning}
\centering
\resizebox{1\linewidth}{!}{
\begin{tabular}{@{}lcccccccccccccccc@{}}
\toprule
\multirow{2}{*}{Method} & \multirow{2}{*}{Optimization}  & \multicolumn{5}{c}{Real Forward-facing~\cite{llff}} & \multicolumn{5}{c}{NeRF Synthetic~\cite{nerf}}& \multicolumn{5}{c}{Tanks and Temples~\cite{tanks}} \\ 
\cmidrule(lr){3-7}\cmidrule(lr){8-12}\cmidrule(lr){13-17}
 & & PSNR $\uparrow$ & SSIM $\uparrow$ & LPIPS $\downarrow$ & Time$_{ft}$ $\downarrow$ & FPS $\uparrow$ & PSNR $\uparrow$ & SSIM $\uparrow$ & LPIPS $\downarrow$ & Time$_{ft}$ $\downarrow$ & FPS $\uparrow$ & PSNR $\uparrow$ & SSIM $\uparrow$ & LPIPS $\downarrow$ & Time$_{ft}$ $\downarrow$ & FPS $\uparrow$ \\
\midrule
NeRF~\cite{nerf} & \multirow{5}{*}{Pipeline} & \underline{25.97} & 0.870 & 0.236 & 10.2h & 0.08 & 30.63 & 0.962 & 0.093 & 10.2h & 0.07 & 21.42 & 0.702 & 0.558 & 10.2h & 0.08 \\
IBRNet~\cite{ibrnet} & & 24.88 & 0.861 & 0.189 & 1.0h & 0.10 & 25.62 & 0.939 & 0.111 & 1.0h & 0.10 & 22.22 & 0.813	& 0.221	& 1.0h & 0.10\\ 
MVSNeRF~\cite{mvsnerf} & & 25.45 & 0.877 & 0.192 & 15min & 0.20 & 27.07 & 0.931 & 0.168 & 15min & 0.19 & 21.83 & 0.841 & 0.235 & 15min & 0.20 \\
ENeRF~\cite{enerf} & & 24.89 & 0.865 & 0.159 & 1.0h & 11.7 & 27.57 & 0.954 & 0.063 & 1.0h & 10.5 & 24.18 & 0.885 & 0.145 & 1.0h & 11.7 \\
Ours & & 25.92 & \underline{0.891} & \underline{0.135} & 1.0h & 14.1 & 27.87 & 0.956 & 0.061 & 1.0h & 12.5 & \underline{24.35} & \underline{0.888} & \textbf{0.125} & 1.0h & 14.0 \\
\midrule
3D-GS$_{7k}$~\cite{3Dgaussians} & \multirow{3}{*}{Gaussians} & 22.15 & 0.808 & 0.243 & 2min & 370 & \underline{32.15} & \underline{0.971} & \underline{0.048} & 1min15s & 450 & 20.13 & 0.778 & 0.319 & 2min30s & 320 \\
3D-GS$_{30k}$~\cite{3Dgaussians} & & 23.92 & 0.822 & 0.213 & 10min & 350 & 31.87 & 0.969 & 0.050 & 7min & 430 & 23.65 & 0.867 & 0.184 & 15min & 270 \\
Ours & & \textbf{26.98} & \textbf{0.913} & \textbf{0.113} & 45s & 350 & \textbf{32.20} & \textbf{0.972} & \textbf{0.043} & 50s & 470 & \textbf{24.58} & \textbf{0.903} & \underline{0.137} & 90s & 330\\
\bottomrule
\end{tabular}}
\end{table}

\begin{figure}[tb]
  \centering
  \includegraphics[width=1\textwidth]{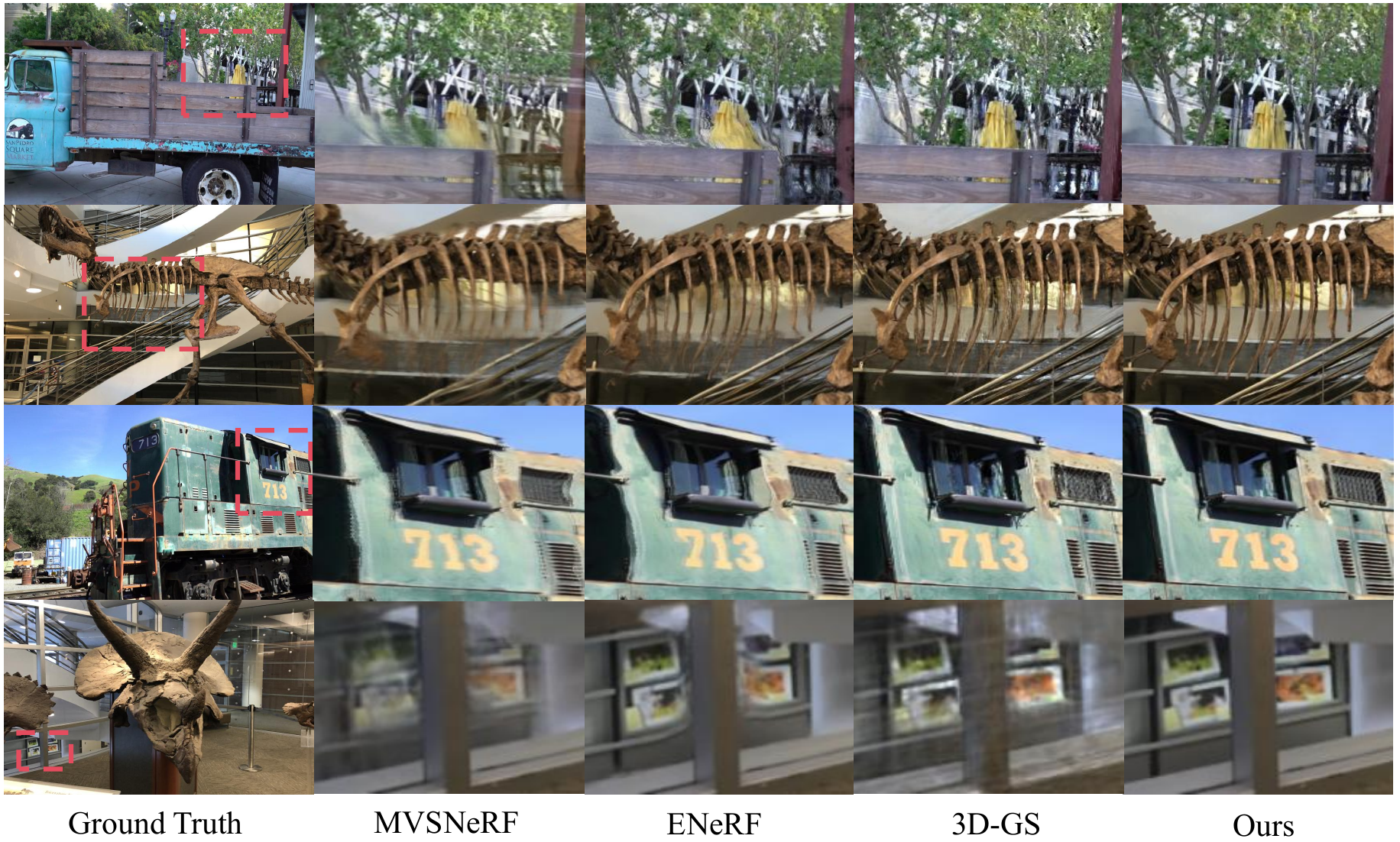}
  \caption{\textbf{Qualitative comparison of rendering quality with state-of-the-art methods~\cite{mvsnerf,enerf,3Dgaussians} after per-scene optimization.} }
  \label{fig:vis_ft}
\end{figure}

\begin{table}[t]
\caption{\textbf{Ablation studies.} The terms ``gs'' and ``vr'' represent Gaussian Splatting and volume rendering, respectively. PSNR$_{dtu}$, PSNR$_{llff}$, PSNR$_{nerf}$, and PSNR$_{tnt}$ are the PSNR metrics for different datasets~\cite{dtu,llff,nerf,tanks}.}
\label{Tab:ablation}
\centering
\setlength{\tabcolsep}{5pt}
\resizebox{0.7\linewidth}{!}{
\begin{tabular}{@{}c|ccc|cccc@{}}
\toprule
& Cascade & Decoding & Color & PSNR$_{dtu}$  & PSNR$_{llff}$ & PSNR$_{nerf}$ & PSNR$_{tnt}$ \\
\midrule
No.1 & \ding{55} & gs & rgb & 26.71 & 22.57 & 24.90 &  21.06 \\
No.2 & \CheckmarkBold & gs & rgb & 27.48 & 23.15 & 25.48 & 21.70\\
No.3 & \CheckmarkBold & vr & rgb & 27.39 & 23.80 & 25.65 & 22.76\\
No.4 & \CheckmarkBold & gs+vr & rgb & 28.21 & 24.07 & 26.46 & 23.28  \\
No.5 & \CheckmarkBold & gs+vr & sh & 28.19 & 23.74 & 24.27 & 22.70\\
\bottomrule
\end{tabular}}
\end{table}

\subsection{Per-Scene Optimization Results}
\label{subsec:per-scene optimization results}
The quantitative results after per-scene optimization are reported in Table~\ref{Tab:fine-tuning}. For per-scene optimization, one strategy is to optimize the entire pipeline, similar to NeRF-based methods. Another approach is to optimize only the initial Gaussian point cloud provided by the generalizable model. When optimizing the entire pipeline, our method can achieve better performance with faster inference speeds compared to previous generalizable NeRF methods, and results comparable to NeRF, demonstrating the robust representation capabilities of our method. In contrast, optimizing only the Gaussians can significantly improve optimization and rendering speed because it eliminates the time-consuming feed-forward neural network. Moreover, performance can benefit from the adaptive density control module described in~\cref{sec:preliminary}. Due to the excellent initialization provided by the generalizable model and the effective aggregation strategy, we achieve optimal performance within a short optimization period, approximately one-tenth of that of 3D-GS. Especially on the Real Forward-facing dataset, our method achieves superior performance with only $45$ seconds of optimization, compared to $10$ minutes for 3D-GS and $10$ hours for NeRF. Additionally, our method's inference speed is comparable to that of 3D-GS and significantly outperforms NeRF-based methods. As shown in Fig.~\ref{fig:vis_ft}, our method is capable of producing high-fidelity views with finer details.

\subsection{Ablations and Analysis}
\label{subsec:ablations and analysis}
\noindent \textbf{Ablation studies.}
As shown in Table~\ref{Tab:ablation}, we conduct ablation studies to evaluate the effectiveness of our designs. Firstly, comparing No.1 and No.2, the cascaded structure demonstrates a significant role. Additionally, adopting the hybrid Gaussian rendering approach (No.4) notably enhances performance compared to utilizing splatting (No.2) or volume rendering (No.3) alone. Regarding color representation, we directly decode RGB values instead of spherical harmonic (SH) coefficients (No.5), as decoding coefficients may result in a degradation of generalization, especially notable on the NeRF Synthetic dataset.

\noindent \textbf{Aggregation strategies.} 
As shown in Table~\ref{Tab:fusion}, we investigate the impact of different point cloud aggregation strategies, which provide varying qualities of initialization and significantly affect subsequent optimization. The direct concatenation approach leads to an excessively large initial point set, hindering optimization and rendering speeds. Downsampling the point cloud can mitigate this issue while also improving performance, as it reduces contamination from noisy points. However, performance remains limited as it also simultaneously reduces some valid points. Employing the consistency check strategy can further boost performance, as it filters out noisy points while preserving valid points.

\begin{figure}[tb]
  \centering
    \begin{subfigure}{0.32\textwidth}
      \centering
      \includegraphics[width=0.95\linewidth]{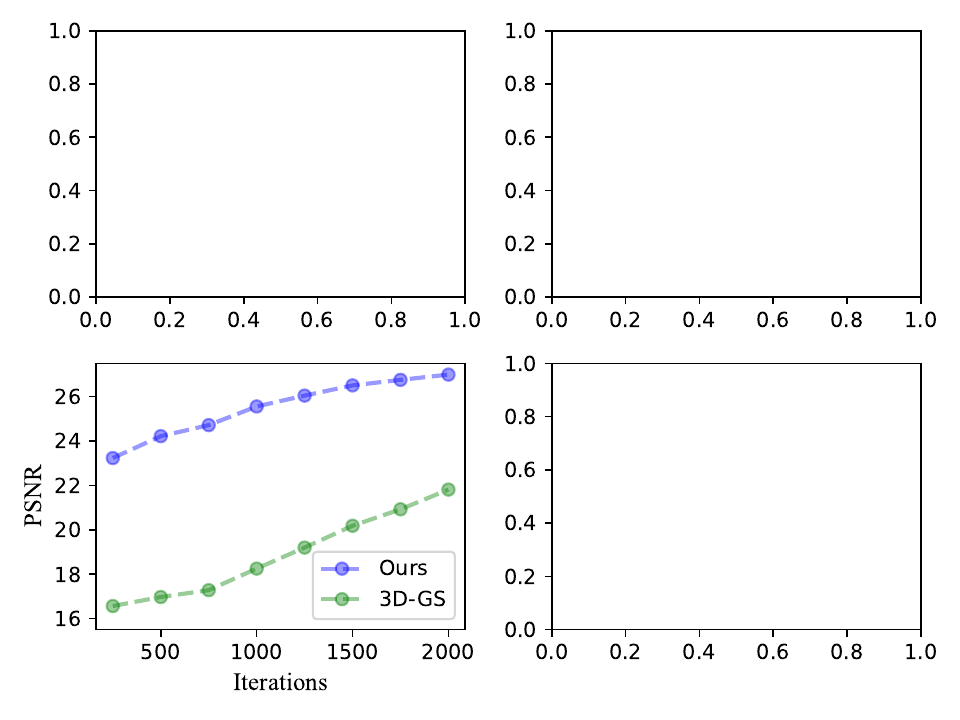}
      \caption{}
    \end{subfigure}
    \hfil
    \begin{subfigure}{0.65\textwidth}
      \centering
      \includegraphics[width=0.9\linewidth]{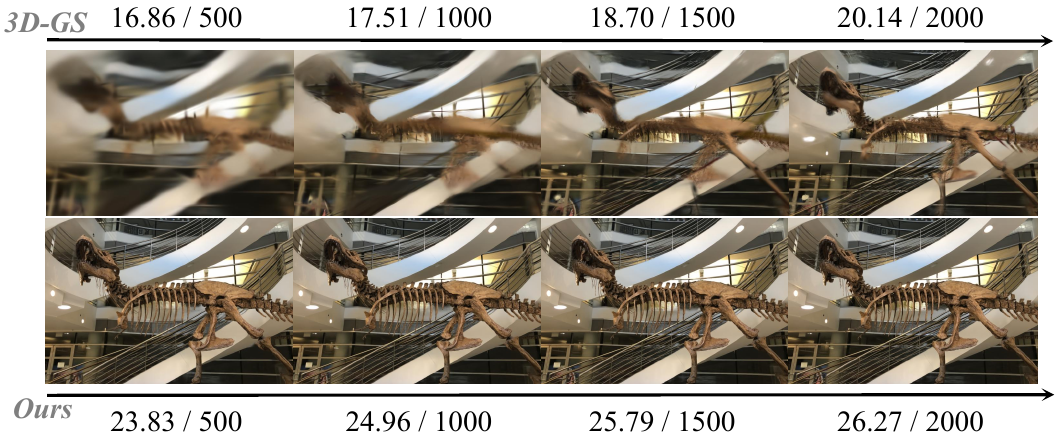}
      \caption{}
    \end{subfigure}
  \caption{\textbf{Analysis of the Optimization process.} (a) The evolution of view quality (PSNR) on the Real Forward-facing~\cite{llff} dataset during the first $2000$ iterations of our method and 3D-GS~\cite{3Dgaussians}. (b) Qualitative comparison of our method (bottom) and 3D-GS (top) on the ``trex'' scene, where (PSNR/iteration number) is shown.}
  \label{fig:ft_process}
\end{figure}

\begin{table}[t]
\caption{\textbf{Comparison of different aggregation strategies.} We report the quantitative results obtained with different strategies on the Real Forward-facing dataset~\cite{llff}. For downsampling aggregation, we employ widely-used voxel downsampling with a voxel size set to $2$. The iteration number for all aggregation strategies is set to $2.5$k.}
\label{Tab:fusion}
\centering
\setlength{\tabcolsep}{5pt}
\resizebox{0.7\linewidth}{!}{
\begin{tabular}{@{}l|ccccc@{}}
\toprule
Aggregation & PSNR $\uparrow$ & SSIM $\uparrow$ & LPIPS $\downarrow$ & Time$_{ft}$ $\downarrow$ & FPS $\uparrow$ \\
\midrule
direct concatenation & 26.18 & 0.901 & 0.122 & 90s & 220 \\
downsampling & 26.72 & 0.909 & 0.121 & 60s & 340 \\
consistency check & 26.98 & 0.913 & 0.113 & 45s & 350 \\
\bottomrule
\end{tabular}}
\end{table}

\noindent \textbf{Optimization process.}
We illustrate the optimization process in Fig.~\ref{fig:ft_process}. Thanks to the excellent initialization provided by the generalizable model, our method quickly attains good performance and rapidly improves. 

\section{Conclusion}
We present MVSGaussian, an efficient generalizable Gaussian Splatting approach. Specifically, we leverage MVS to infer depth, establishing a pixel-aligned Gaussian representation. To enhance generalization, we propose a hybrid rendering approach that integrates depth-aware volume rendering. Besides, thanks to high-quality initialization, our models can be fine-tuned quickly for specific scenes. Compared with generalizable NeRFs, which typically require minutes of fine-tuning and seconds of rendering per image, MVSGaussian achieves real-time rendering with superior synthesis quality. Moreover, compared with 3D-GS, MVSGaussian achieves better view synthesis with reduced training time. 

\noindent \textbf{Limitations.}
As our method relies on MVS for depth estimation, it inherits limitations from MVS, such as decreased depth accuracy in areas with weak textures or specular reflections, resulting in degraded view quality.

%
%
\bibliographystyle{splncs04}
\bibliography{main}

\clearpage

\appendix
\section{Implementation and Network Details}

\begin{algorithm}[t]
    \SetAlgoLined
    \KwIn{Camera parameters, Depth maps $D_0$ and $\{D_i\}_{i=1}^{N}$, predefined thresholds $\{\theta_p(n)\}_{n=1}^{N_{\theta}}$ and $\{\theta_d(n)\}_{n=1}^{N_{\theta}}$}
    \KwOut{$Mask$}
    \textbf{Initialization:} $ Mask \gets 0 $ \\
    \For {$i$ \bf{in} ($1,...,N$)}
        {
            $Err_p^{i} \gets zeros(H,W), Err_d^{i} \gets zeros(H,W)$ \\
            \For {$p$ \bf{in} $(0,0)$ to $(H-1,W-1)$}
            {
                $\xi_p^{i} \gets \left\|{p - p'}\right\|_2$, \Comment{calculate the reprojetcion error between $D_0$ and $D_i$} \\
                $\xi_d^{i} \gets {\left\|{D_0(p) - d'}\right\|_1}/{D_0(p)}$\\
                $Err_p^{i}(p) \gets \xi_p^{i}$ \\
                $Err_d^{i}(p) \gets \xi_d^{i}$
            }
        \For {$n$ \bf{in} ($1,...,N_{\theta}$)}
            {
                $ Mask^i_n \gets (Err_p^{i} < \theta_p(n)) \& (Err_d^{i} < \theta_d(n)$)
            }
        }
    \For {$n$ \bf{in} ($1,...,N_{\theta}$)}
    {
        $Mask_n \gets 0$ \\ 
        \For {$i$ \bf{in} ($1,...,N$)}
        {
            $Mask_n \gets Mask_n + Mask^i_n$
        }
        $Mask_n \gets (Mask_n>n)$ \\
        $Mask \gets Mask \cup Mask_n $
    }
    \caption{Dynamic Consistency Checking}
    \label{algorithm 1}
\end{algorithm} 

\begin{figure}[t]
  \centering
  \includegraphics[width=0.8\textwidth]{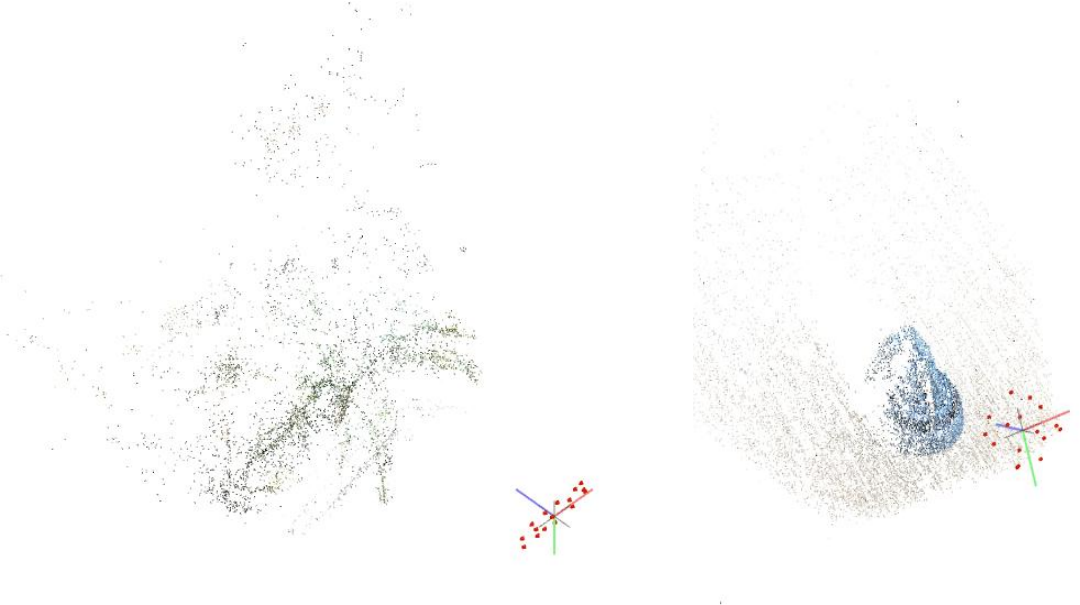}
  \caption{\textbf{Visualization of camera calibration and point cloud reconstruction by COLMAP.} }
  \label{fig:supp_colmap_vis}
\end{figure}

\noindent \textbf{Implementation Details.}
Following ENeRF~\cite{enerf}, we partition the DTU~\cite{dtu} dataset into $88$ training scenes and $16$ test scenes. We train the generalizable model on four RTX $3090$ GPUs using the Adam~\cite{adam} optimizer, with an initial learning rate set to $5e-4$. The learning rate is halved every $50$k iterations. During the training process, we select $2$, $3$, and $4$ source views as inputs with respective probabilities of $0.1$, $0.8$, and $0.1$. For evaluation, we follow the criteria established in prior works such as ENeRF~\cite{enerf} and MVSNeRF~\cite{mvsnerf}. Specifically, for the DTU test set, segmentation masks are employed to evaluate performance, defined based on the availability of ground-truth depth at each pixel. For Real Forward-facing dataset~\cite{llff}, where the marginal region of images is typically invisible to input images, we evaluate the $80\%$ area in the center of the images. This evaluation methodology is also applied to the Tanks and Temples dataset~\cite{tanks}. The image resolutions of the DTU, the Real
Forward-facing, the NeRF Synthetic~\cite{nerf}, and the Tanks and Temples datasets are $512\times640$, $640\times960$, $800\times800$, and $640\times960$ respectively. As discussed in Sec.~4.3 of the main text, we employ a consistency check to filter out noisy points for high-quality initialization. Specifically, we apply a dynamic consistency checking algorithm~\cite{D2HC-RMVSNet,et-mvsnet}, the details of which are provided in Algorithm~\ref{algorithm 1}. The predefined thresholds $\{\theta_p(n)\}_{n=1}^{N_{\theta}}$ are set to $\{\frac{n}{8}\}_{n=1}^{N_{\theta}}$, and $\{\theta_d(n)\}_{n=1}^{N_{\theta}}$ are set to $\{\frac{n}{10}\}_{n=1}^{N_{\theta}}$. For 3D-GS~\cite{3Dgaussians}, following~\cite{FSGS}, we use COLMAP~\cite{colmap} to reconstruct the point cloud from the working set (training views) as initialization. Specifically, we employ COLMAP's automatic reconstruction to achieve the reconstruction of sparse point clouds. Some examples are shown in Fig.~\ref{fig:supp_colmap_vis}. As mentioned in Sec.5.1 of the main text, our optimization strategy and hyperparameters settings remain consistent with the vanilla 3D-GS, except for the number of iterations. The iterations of our method on Real Forward-facing, NeRF Synthetic and Tanks and Temples datasets are $2.5$k, $5$k and $5$k, respectively.

\noindent \textbf{Network Details.}
As mentioned in Sec.~4.2 of the main text, we apply a pooling network $\rho$ to aggregate multi-view features to obtain the aggregated features via $f_v = \rho(\{f_i\}_{i=1}^N)$. The implementation details are consistent with~\cite{enerf}: initially, the mean $\mu$ and variance $v$ of $\{f_i\}_{i=1}^N$ are computed. Subsequently, $\mu$ and $v$ are concatenated with each $f_i$ and an MLP is applied to generate weights. The $f_v$ is then blended using a soft-argmax operator, combining the obtained weights and multi-view features $(\{f_i\}_{i=1}^N)$.

\begin{table}[t]
\caption{\textbf{The performance of our method with varying numbers of input views on the DTU, Real Forward-facing, and NeRF Synthetic datasets.} ``Mem'' and ``FPS'' are measured under the image resolution of $512\times640$.}
\label{tab:supp_ab_view}
\centering
\resizebox{1\linewidth}{!}{
\begin{tabular}{@{}cccccccccccc@{}}
\toprule
\multirow{2}{*}{Views} & \multicolumn{3}{c}{DTU~\cite{dtu}} & \multicolumn{3}{c}{Real Forward-facing~\cite{llff}} & \multicolumn{3}{c}{NeRF Synthetic~\cite{nerf}} & \multirow{2}{*}{Mem(GB)$\downarrow$} & \multirow{2}{*}{FPS$\uparrow$}\\ 
\cmidrule(lr){2-4}\cmidrule(lr){5-7}\cmidrule(lr){8-10}
 & PSNR $\uparrow$ & SSIM $\uparrow$ & LPIPS $\downarrow$ & PSNR $\uparrow$ & SSIM $\uparrow$ & LPIPS $\downarrow$ & PSNR $\uparrow$ & SSIM $\uparrow$ & LPIPS $\downarrow$ \\
\midrule
2 & 25.78 & 0.947 & 0.095 & 23.11 & 0.834 & 0.175 & 25.06 & 0.937 & 0.079 & 0.866 & 24.5 \\
3 & 28.21 & 0.963 & 0.076 & 24.07 & 0.857 & 0.164 & 26.46 & 0.948 & 0.071 &  0.876 & 21.5 \\
4 & 28.43 & 0.965 & 0.075 & 24.46 & 0.870 & 0.164 & 26.50 & 0.949 & 0.071 & 1.106 & 19.1 \\
\bottomrule
\end{tabular}}
\end{table}

\section{Additional Ablation Experiments}
\noindent \textbf{Numbers of Views.}
Existing generalizable Gaussian methods, such as PixelSplat~\cite{pixelsplat} and GPS-Gaussian~\cite{gpsgaussian}, focus on image pairs as input, while Splatter Image~\cite{splatterimage} prioritize single-view reconstruction. Our method is view-agnostic, capable of supporting varying numbers of views as input. We report the performance with varying numbers of input views in Table~\ref{tab:supp_ab_view}. As the number of views increases, the model can leverage more scene information, leading to improved performance. Meanwhile, increasing the number of views only introduces a slight increase in computational cost and memory consumption.

\noindent \textbf{Numbers of Sampled Points.}
In the main text, we apply a pixel-align Gaussian representation, where each pixel is unprojected into 3D space based on the estimated depth, corresponding to a 3D Gaussian. An alternative approach is to sample $M$ depths centered at the estimated depth map, resulting in each pixel being unprojected into $M$ Gaussians. As shown in Table~\ref{tab:supp_ab_sampled_points}, increasing the number of 3D sampled points improves performance but raises computational costs.  To strike a balance between cost and performance, we set $M=1$.

\begin{table}[t]
\caption{\textbf{The performance of different numbers of sampled points on the DTU, Real Forward-facing, NeRF Synthetic, and Tanks and Temples datasets.} Here, ``Samples" represents the number of 3D points sampled along the ray for each pixel.}
\label{tab:supp_ab_sampled_points}
\centering
\resizebox{1\linewidth}{!}{
\begin{tabular}{@{}ccccccccccccccc@{}}
\toprule
\multirow{2}{*}{Samples} & \multicolumn{3}{c}{DTU~\cite{dtu}} & \multicolumn{3}{c}{Real Forward-facing~\cite{llff}} & \multicolumn{3}{c}{NeRF Synthetic~\cite{nerf}} & \multicolumn{3}{c}{Tanks and Temples~\cite{tanks}} & \multirow{2}{*}{Mem(GB)$\downarrow$} & \multirow{2}{*}{FPS$\uparrow$}\\ 
\cmidrule(lr){2-4}\cmidrule(lr){5-7}\cmidrule(lr){8-10}\cmidrule(lr){11-13}
 & PSNR $\uparrow$ & SSIM $\uparrow$ & LPIPS $\downarrow$ & PSNR $\uparrow$ & SSIM $\uparrow$ & LPIPS $\downarrow$ & PSNR $\uparrow$ & SSIM $\uparrow$ & LPIPS $\downarrow$ & PSNR $\uparrow$ & SSIM $\uparrow$ & LPIPS $\downarrow$ \\
\midrule
1 & 28.21 & 0.963 & 0.076 & 24.07 & 0.857 & 0.164 & 26.46 & 0.948 & 0.071 & 23.28 & 0.877 & 0.139 & 0.876 & 21.5 \\
2 & 28.26 & 0.963 & 0.075 & 24.20 & 0.861 & 0.163 & 26.64 & 0.949 & 0.070 & 23.20 & 0.879 & 0.151 & 1.508 & 19.0  \\
\bottomrule
\end{tabular}}
\end{table}

\noindent \textbf{Density for Volume Rendering.}
Since only one point per ray is sampled, our model predicts single radiance $r$ and density $\sigma$. In this case, the pixel's color $c$ obtained through volume rendering is given by $c=(1-\exp(-\sigma))r$. This resembles pixel-aligned splatting, where one Gaussian contributes one pixel, sharing the alpha-based rendering principles but offering a simpler implementation. Therefore, predicting density is necessary as it indicates the point's opacity, as validated by ablation results in Table~\ref{tab:supp_density}.

\begin{table}[t]
\caption{\textbf{The ablation study on the density prediction.}}
\label{tab:supp_density}
\centering
\resizebox{0.75\linewidth}{!}{
\begin{tabular}{@{}cccccccccc@{}}
\toprule
\multirow{2}{*}{Settings} & \multicolumn{3}{c}{DTU} & \multicolumn{3}{c}{Real Forward-facing} & \multicolumn{3}{c}{NeRF Synthetic}\\ 
\cmidrule(lr){2-4}\cmidrule(lr){5-7}\cmidrule(lr){8-10}
 & PSNR  & SSIM  & LPIPS  & PSNR & SSIM  & LPIPS  & PSNR  & SSIM & LPIPS \\
\midrule
w/o density & 28.03 & 0.963 & 0.076 & 23.96 & 0.854 & 0.165 & 26.22 & 0.947 & 0.071 \\
w density & 28.21 & 0.963 & 0.076 & 24.07 & 0.857 & 0.164 & 26.46 & 0.948 & 0.071 \\
\bottomrule
\end{tabular}}
\end{table}

\noindent \textbf{Initialization Comparison.}
Our generalization model can provide a denser point cloud for 3D-GS~\cite{3Dgaussians} than Structure-from-Motion (SfM) as initialization. Considering that the MVS method can also obtain a denser point cloud, we conduct the comparison in Table~\ref{tab:supp_init}. MVS methods are typically trained on DTU~\cite{dtu} and BlendedMVS~\cite{yao2020blendedmvs}, then tested on Tanks and Temples dataset~\cite{tanks}. Thus, we select the latest ET-MVSNet~\cite{et-mvsnet} and compare it on Tanks and Temples dataset. While ET-MVSNet~\cite{et-mvsnet} surpasses SfM, it's still limited. Because it focuses solely on accurate depth, while our method generates point clouds tailored for view synthesis. Depth and view quality aren't directly proportional, as mentioned in previous works such as ENeRF~\cite{enerf}.

\begin{table}[t]
\caption{\textbf{Initialization Comparison.}}
\label{tab:supp_init}
\centering
\resizebox{0.5\linewidth}{!}{
\begin{tabular}{@{}l|ccccc@{}}
\toprule
Initialization & PSNR & SSIM & LPIPS & Time$_{ft}$ & FPS \\
\midrule
ET-MVSNet~\cite{et-mvsnet} & 22.66 & 0.861 & 0.204 & 90s & 300$+$ \\
Ours & 24.58 & 0.903 & 0.137 & 90s & 300$+$ \\
\bottomrule
\end{tabular}}
\end{table}

\begin{figure}
  \centering
  \includegraphics[width=0.95\textwidth]{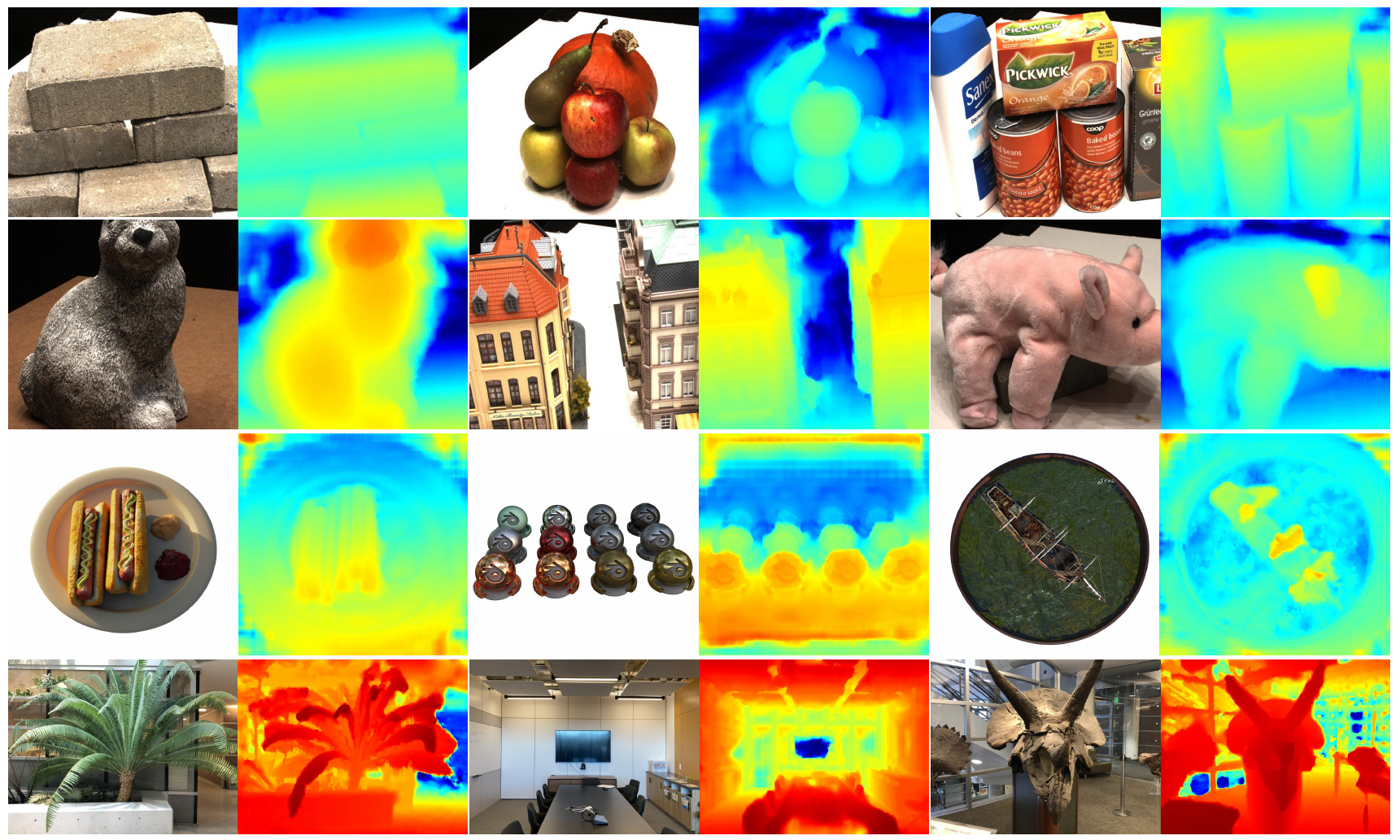}
  \caption{\textbf{Depth maps visualization.} We visualize the depth maps predicted by our method on different datasets~\cite{dtu,nerf,llff}.}
  \label{fig:supp_depth_vis}
\end{figure}

\noindent \textbf{Depth Analysis.}
Benefiting from the explicit geometry reasoning of MVS, our method can produce reasonable depth maps, as illustrated in Fig.~\ref{fig:supp_depth_vis}. The quantitative results are shown in Table~\ref{tab:supp_depth}. Compared with previous generalizable NeRF methods, our method can achieve the most accurate depth estimation.

\begin{table}[t]
\caption{\textbf{Quantitative comparison of depth reconstruction on the DTU test set.} MVSNet is trained using depth supervision, while other methods are trained with only RGB image supervision. ``Abs err'' refers to the average absolute error, and ``Acc(X)'' denotes the percentage of pixels with an error less than X mm.}
\label{tab:supp_depth}
\centering
\scalebox{1}{
\resizebox{0.65\linewidth}{!}{
\begin{tabular}{@{}lcccccc@{}}
\toprule
\multirow{2}{*}{Method} & \multicolumn{3}{c}{Reference view} & \multicolumn{3}{c}{Novel view} \\ 
\cmidrule(lr){2-4}\cmidrule(lr){5-7}
& Abs err $\downarrow$ & Acc(2)$\uparrow$ & Acc(10)$\uparrow$ & Abs err $\downarrow$ & Acc(2)$\uparrow$ & Acc(10)$\uparrow$ \\
\midrule
MVSNet~\cite{mvsnet} & 3.60 & 0.603 & 0.955 & - & - & - \\
PixelNeRF~\cite{pixelnerf} & 49 & 0.037 & 0.176 & 47.8 & 0.039 & 0.187 \\
IBRNet~\cite{ibrnet} & 338 & 0.000 & 0.913 & 324 & 0.000 & 0.866  \\
MVSNeRF~\cite{mvsnerf} & 4.60 & 0.746 & 0.913 & 7.00 & 0.717 & 0.866 \\
ENeRF-MVS~\cite{enerf} & 3.80 & 0.823 & 0.937 & 4.80 & 0.778 & 0.915 \\
ENeRF-NeRF~\cite{enerf} & 3.80 & 0.837 & 0.939 & 4.60 & 0.792 & 0.917 \\
Ours & \textbf{3.11} & \textbf{0.866} & \textbf{0.956} & \textbf{3.66} & \textbf{0.838} & \textbf{0.945} \\
\bottomrule
\end{tabular}}
}
\end{table}

\noindent \textbf{Point Cloud Analysis.}
As discussed in Sec.~5.4 of the main text, different point cloud aggregation strategies can provide varying-quality initialization for subsequent per-scene optimization. Here, we report the initial and final numbers of point clouds in Table~\ref{tab:supp_point_cloud} and provide the visual comparison in Fig.~\ref{fig:supp_point_vis}. The direct concatenation approach leads to excessively large initialization point clouds, which slow down optimization and rendering speeds. The down-sampling approach can reduce the total number of points and mitigate noisy points, but it also leads to a reduction in effective points. Our applied consistency check strategy can filter out noisy points while retaining effective ones.

\begin{table}[t]
\caption{\textbf{Comparison of point cloud quantities under different aggregation strategies on the Real Forward-facing dataset.} For downsampling, we employ widely-used voxel downsampling, with a voxel size set to $2$. The iteration number for all strategies is set to $2.5$k.}
\label{tab:supp_point_cloud}
\centering
\scalebox{1}{
\resizebox{0.6\linewidth}{!}{
\begin{tabular}{@{}lcc@{}}
\toprule
Strategy & initial points(k) & final points(k) \\ 
\midrule
direct concatenation & 2458 & 2176 \\
downsampling  & 836 & 839 \\
consistency check & 860 & 913 \\
\bottomrule
\end{tabular}}
}
\end{table}

\begin{table}[t]
\caption{\textbf{Time overhead for each module (in milliseconds).}}
\label{tab:supp_time}
\centering
\resizebox{0.55\linewidth}{!}{
\begin{tabular}{@{}l|cccc@{}}
\toprule
Modules & coarse stage & & & fine stage \\
\midrule
Feature extractor &  \multicolumn{4}{c}{1.3} \\
\midrule
Depth estimation & 8.1 & & & 7.9 \\
Gaussian representation & - & & & 24.0 \\
Gaussian rendering & - & & & 4.4 \\
\bottomrule
\end{tabular}
}
\end{table}

\noindent \textbf{Inference Speed Analysis.}
As shown in Table~1 of the main text, the inference speed (FPS) of our generalizable model is $21.5$. Here, we present the inference time breakdown result in Table~\ref{tab:supp_time}. The primary time overhead comes from the neural network, while the subsequent rendering process incurs minimal time overhead. Therefore, we discard the neural network component during the per-scene optimization stage, resulting in a significant increase in speed.

\section{More Qualitative Results}
\begin{figure}[t]
  \centering
  \includegraphics[width=0.95\textwidth]{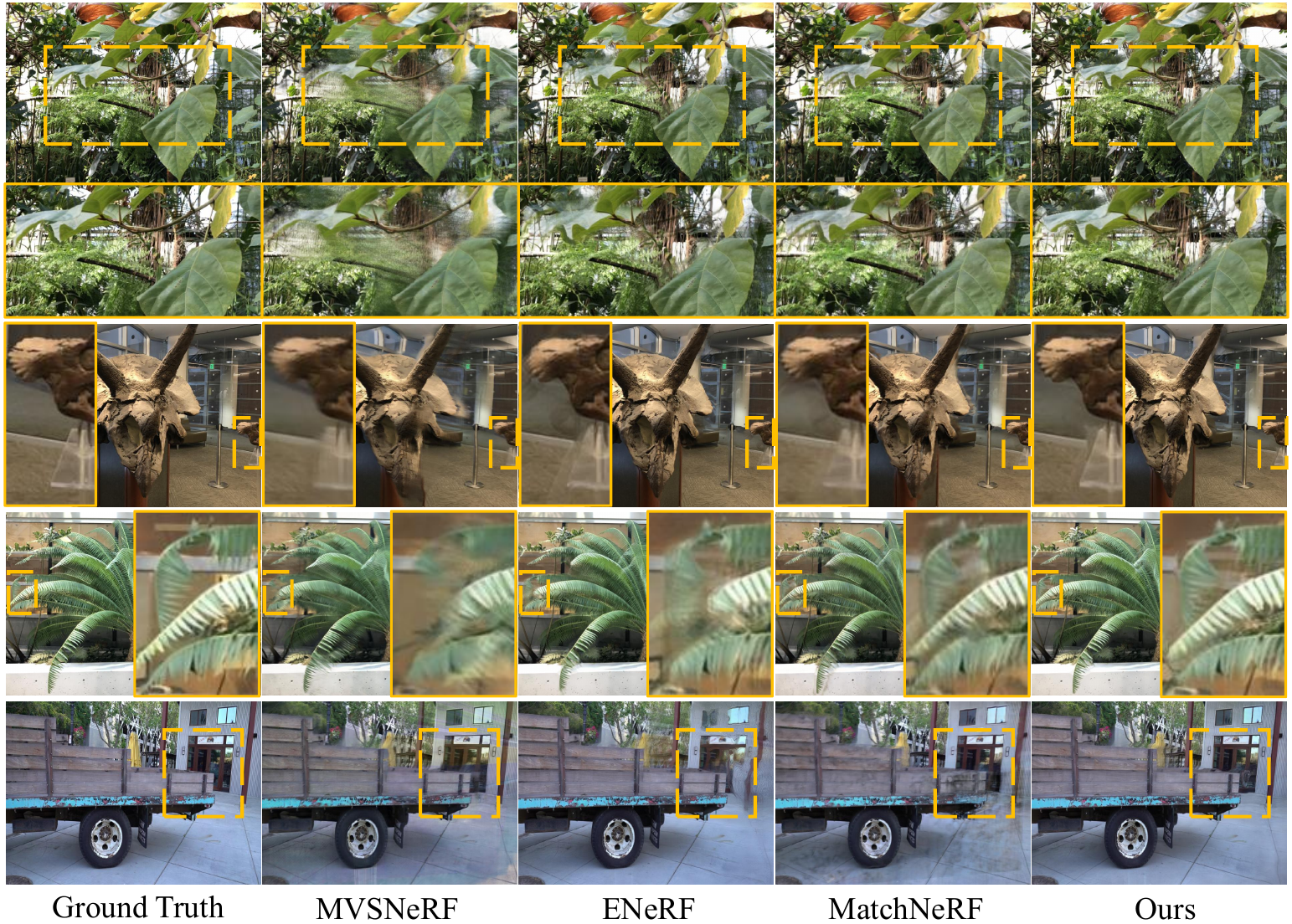}
  \caption{\textbf{Qualitative comparison of rendering quality with state-of-the-art methods~\cite{mvsnerf,matchnerf,enerf} under generalization and three views settings.}}
  \label{fig:supp_gen_vis}
\end{figure}

\begin{figure}
  \centering
  \includegraphics[width=1\textwidth]{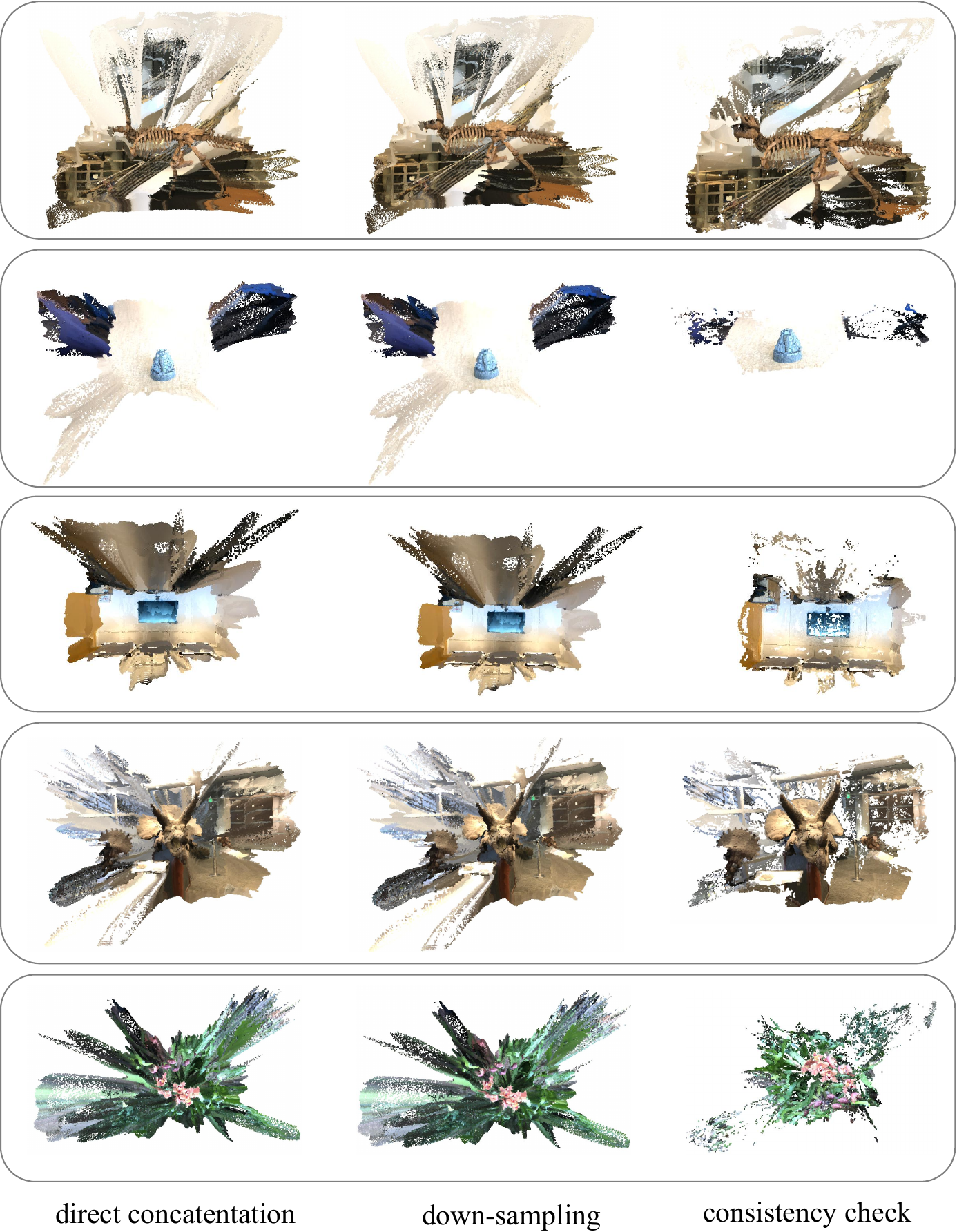}
  \caption{\textbf{Point cloud visualization under different aggregation strategies.} }
  \label{fig:supp_point_vis}
\end{figure}

\noindent \textbf{Qualitative Results under the Generalization Setting.}
As shown in Fig.~\ref{fig:supp_gen_vis}, we present qualitative comparisons of the generalization results obtained by different methods. Our method is capable of producing higher-fidelity views, particularly in some challenging areas. For instance, in geometrically complex scenes, around objects' edges, and in reflective areas, our method can reconstruct more details while exhibiting fewer artifacts.

\begin{figure}
  \centering
  \includegraphics[width=1\textwidth]{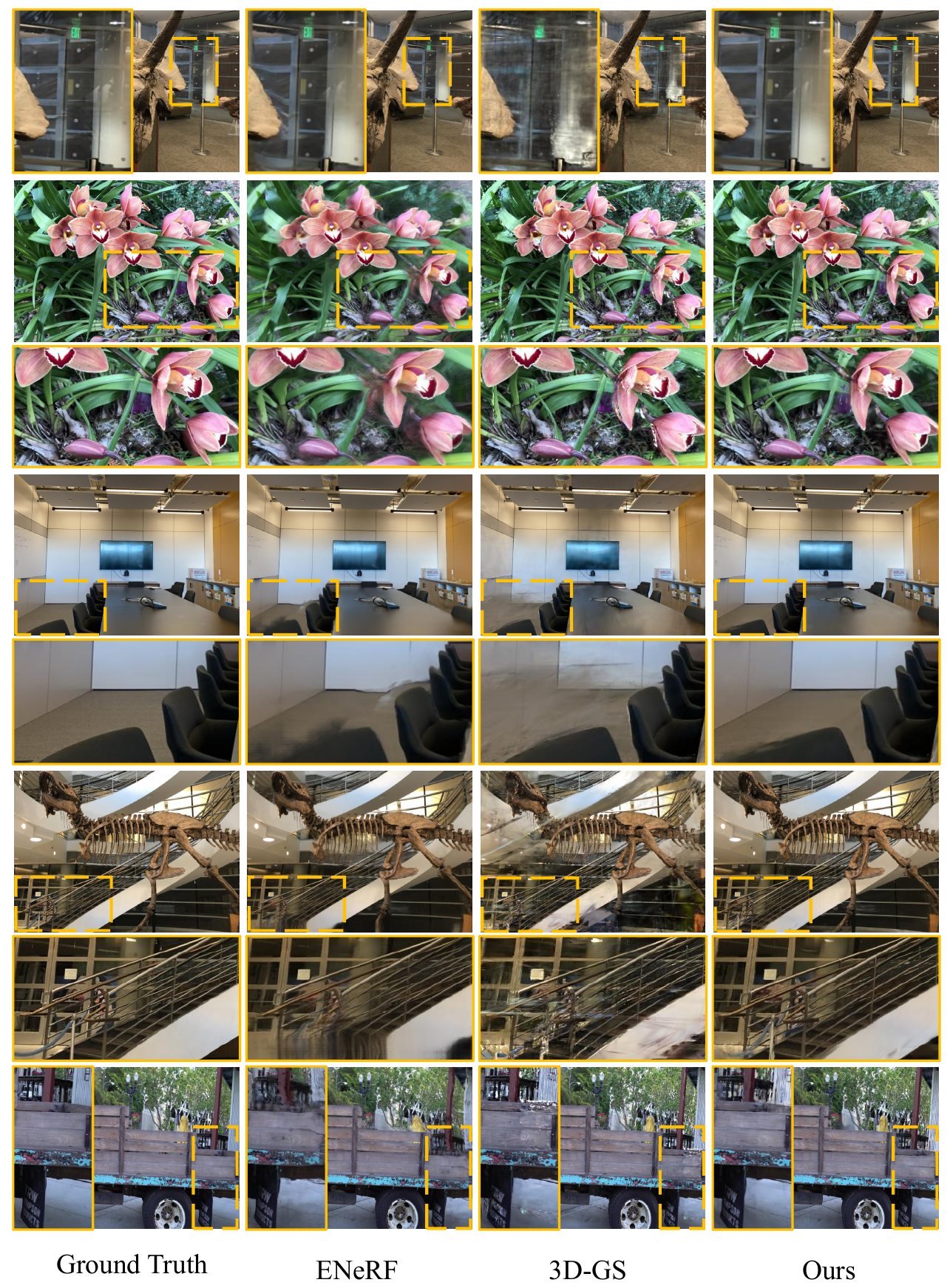}
  \caption{\textbf{Qualitative comparison of rendering quality with state-of-the-art methods~\cite{enerf,3Dgaussians} after per-scene optimization.}}
  \label{fig:supp_ft_vis}
\end{figure}

\noindent \textbf{Qualitative Results under the Per-scene Optimization Setting.}
As shown in Fig.~\ref{fig:supp_ft_vis}, we present the visual comparison after fine-tuning. 
Benefiting from the strong initialization provided by our generalizable model, excellent performance can be achieved with just a short fine-tuning period. The views synthesized by our method preserve more scene details and exhibit fewer artifacts.

\section{Per-scene Breakdown}
As shown in Tables~\ref{tab:dtu_break},~\ref{tab:nerf_break},~\ref{tab:llff_break}, and~\ref{tab:tnt_break}, we present the per-scene breakdown results of DTU~\cite{dtu}, NeRF Synthetic~\cite{nerf}, Real Forward-facing~\cite{llff}, and Tanks and Temples~\cite{tanks} datasets. These results align with the averaged results presented in the main text.

\begin{table}
\caption{\textbf{Quantitative per-scene breakdown results on the DTU test set.} PixelSplat$^*$ and Ours$^*$ represent the results obtained with a 2-view input, while the others are the results obtained with a 3-view input.}
\label{tab:dtu_break}
\centering
\resizebox{\linewidth}{!}{
\begin{tabular}{@{}l|cccccccccccccccc@{}}
\toprule
Scan & \#1 & \#8 & \#21 & \#103 & \#114 & \#30 & \#31 & \#34 & \#38 & \#40 & \#41 & \#45 & \#55 & \#63 & \#82 & \#110\\
\midrule
Metric & \multicolumn{16}{c}{PSNR $\uparrow$} \\
\midrule
PixelNeRF~\cite{pixelnerf} & 21.64 & 23.70 & 16.04 & 16.76 & 18.40 & - & - & - & - & - & - & - & - & - & - & - \\
IBRNet~\cite{ibrnet} & 25.97 & 27.45 & 20.94 & 27.91 & 27.91 & - & - & - & - & - & - & - & - & - & - & - \\
MVSNeRF~\cite{mvsnerf} & 26.96 & 27.43 & 21.55 & 29.25 & 27.99 & - & - & - & - & - & - & - & - & - & - & - \\
ENeRF~\cite{enerf} & \underline{28.85} & \underline{29.05} & 22.53 & \underline{30.51} & \underline{28.86} & \underline{29.20} & \underline{25.13} & \underline{26.77} & \underline{28.61} & \underline{25.67} & \underline{29.51} & \textbf{24.83} & \underline{30.26} & 27.22 & 26.83 & \underline{27.97} \\
MatchNeRF~\cite{matchnerf} & 27.69 & 27.76 & \underline{22.75} & 29.35 & 28.16 & 29.16 & 24.26 & 25.66 & 27.52 & 25.16 & 28.27 & 23.94 & 26.64 & \textbf{29.40} & \underline{27.65} & 27.15 \\
Ours & \textbf{29.67} & \textbf{29.65} & \textbf{23.24} & \textbf{30.60} & \textbf{29.26} & \textbf{30.10} & \textbf{25.94} & \textbf{26.82} & \textbf{29.27} & \textbf{26.13} & \textbf{30.33} & \underline{24.55} & \textbf{31.40} & \underline{28.46} & \textbf{27.82} & \textbf{28.15} \\
\midrule
PixelSplat$^{*}$~\cite{pixelsplat} & 14.65 & 14.72 & 10.69 & 16.88 & 15.31 & 10.93 & 13.28 & 14.70 & 14.81 & 13.26 & 16.09 & 12.62 & 15.76 & 12.18 & 12.11 & 16.18 \\
Ours$^{*}$ & 27.22 & 26.88 & 20.49 & 28.25 & 27.89 & 27.55 & 22.96 & 25.32 & 27.13 & 22.89 & 27.71 & 21.78 & 28.85 & 27.01 & 24.64 & 25.92 \\
\midrule
Metric & \multicolumn{16}{c}{SSIM $\uparrow$} \\
\midrule
PixelNeRF~\cite{pixelnerf} & 0.827 & 0.829 & 0.691 & 0.836 & 0.763 & - & - & - & - & - & - & - & - & - & - & -\\
IBRNet~\cite{ibrnet} & 0.918 & 0.903 & 0.873 & 0.950 & 0.943 & - & - & - & - & - & - & - & - & - & - & -\\
MVSNeRF~\cite{mvsnerf} & 0.937 & 0.922 & 0.890 & 0.962 & 0.949 & - & - & - & - & - & - & - & - & - & - & -\\
ENeRF~\cite{enerf} & \underline{0.958} & \underline{0.955} & \underline{0.916} & \underline{0.968} & \underline{0.961} & \underline{0.981} & \underline{0.937} & \underline{0.934} & \underline{0.946} & \underline{0.947} & \underline{0.960} & \underline{0.948} & \underline{0.973} & \underline{0.978} & \underline{0.971} & \underline{0.974} \\
MatchNeRF~\cite{matchnerf} & 0.936 & 0.918 & 0.901 & 0.961 & 0.948 & 0.974 & 0.921 & 0.874 & 0.902 & 0.903 & 0.936 & 0.934 & 0.929 & 0.976 & 0.966 & 0.962 \\
Ours & \textbf{0.966} & \textbf{0.961} & \textbf{0.930} & \textbf{0.970} & \textbf{0.963} & \textbf{0.983} & \textbf{0.946} & \textbf{0.947} & \textbf{0.954} & \textbf{0.957} & \textbf{0.967} & \textbf{0.954} & \textbf{0.979} & \textbf{0.980} & \textbf{0.974} & \textbf{0.976} \\
\midrule
PixelSplat$^{*}$~\cite{pixelsplat} & 0.690 & 0.706 & 0.492 & 0.778 & 0.651 & 0.782 & 0.624 & 0.534 & 0.513 & 0.571 & 0.714 & 0.541 & 0.624 & 0.807 & 0.769 & 0.802 \\
Ours$^{*}$ & 0.950 & 0.948 & 0.895 & 0.963 & 0.954 & 0.977 & 0.919 & 0.925 & 0.933 & 0.928 & 0.951 & 0.933 & 0.967 & 0.974 & 0.965 & 0.966 \\
\midrule
Metric & \multicolumn{16}{c}{LPIPS $\downarrow$} \\
\midrule
PixelNeRF~\cite{pixelnerf} & 0.373 & 0.384 & 0.407 & 0.376 & 0.372 & - & - & - & - & - & - & - & - & - & - & -\\
IBRNet~\cite{ibrnet} & 0.190 & 0.252 & 0.179 & 0.195 & 0.136 & - & - & - & - & - & - & - & - & - & - & -\\
MVSNeRF~\cite{mvsnerf} & 0.155 & 0.220 & 0.166 & 0.165 & 0.135 & - & - & - & - & - & - & - & - & - & - & -\\
ENeRF~\cite{enerf} & \underline{0.086} & \underline{0.119} & \underline{0.107} & \underline{0.107} & \underline{0.076} & \underline{0.052} & \underline{0.108} & \underline{0.117} & \underline{0.118} & \underline{0.120} & \underline{0.091} & \underline{0.077} & \underline{0.069} & \underline{0.048} & \underline{0.066} & \underline{0.069} \\
MatchNeRF~\cite{matchnerf} & 0.157 & 0.227 & 0.149 & 0.179 & 0.132 & 0.085 & 0.169 & 0.234 & 0.220 & 0.216 & 0.174 & 0.127 & 0.164 & 0.077 & 0.093 & 0.141 \\
Ours & \textbf{0.069} & \textbf{0.102} & \textbf{0.088} & \textbf{0.098} & \textbf{0.070} & \textbf{0.048} & \textbf{0.093} & \textbf{0.097} & \textbf{0.098} & \textbf{0.101} & \textbf{0.075} & \textbf{0.067} & \textbf{0.055} & \textbf{0.041} & \textbf{0.057} & \textbf{0.057} \\
\midrule
PixelSplat$^{*}$~\cite{pixelsplat} & 0.423 & 0.366 & 0.471 & 0.357 & 0.366 & 0.329 & 0.429 & 0.435 & 0.493 & 0.427 & 0.438 & 0.488 & 0.343 & 0.278 & 0.326 & 0.254 \\
Ours$^{*}$ & 0.087 & 0.118 & 0.121 & 0.114 & 0.079 & 0.057 & 0.126 & 0.118 & 0.126 & 0.132 & 0.093 & 0.090 & 0.074 & 0.049 & 0.067 & 0.079 \\
\bottomrule
\end{tabular}}
\end{table}

\begin{table}
\caption{\textbf{Quantitative per-scene breakdown results on the Tanks and Temples dataset.} PixelSplat$^*$ and Ours$^*$ represent the results obtained with a 2-view input and low-resolution images, while the other generalizable results are obtained with a 3-view input.}
\label{tab:tnt_break}
\centering
\renewcommand\arraystretch {1}
\resizebox{0.7\linewidth}{!}{
\begin{tabular}{@{}lcccccccc@{}}
\toprule
\multirow{2}{*}{Scene} & \multicolumn{3}{c}{Train} & \multicolumn{3}{c}{Truck}\\ 
\cmidrule(lr){2-4}\cmidrule(lr){5-7}
& PSNR $\uparrow$ & SSIM $\uparrow$ & LPIPS $\downarrow$ & PSNR $\uparrow$ & SSIM $\uparrow$ & LPIPS $\downarrow$ \\
\midrule
IBRNet~\cite{ibrnet} & 22.35 & 0.763 & 0.285 & 19.13 & 0.755 & 0.280 \\
MVSNeRF~\cite{mvsnerf} & 20.58 & 0.816 & 0.278 & 21.16 & 0.830 & 0.242 \\
ENeRF~\cite{enerf} & \underline{22.54} & \underline{0.851} & \underline{0.204} & \underline{22.53} & \underline{0.856} & \underline{0.163} \\
MatchNeRF~\cite{matchnerf} & 20.44 & 0.789 & 0.332 & 21.16 & 0.796 & 0.269 \\
Ours & \textbf{23.00} & \textbf{0.872} & \textbf{0.154} & \textbf{23.55} & \textbf{0.883} & \textbf{0.124} \\
\midrule
PixelSplat$^{*}$~\cite{pixelsplat} & 18.21 & 0.638 & 0.252 & 20.58 & 0.741 & 0.195 \\
Ours$^{*}$ & 23.67 & 0.864 & 0.132 & 22.68 & 0.834 & 0.127 \\
\midrule
NeRF~\cite{nerf} & 21.02 & 0.707 & 0.538 & 21.82 & 0.696 & 0.577 \\ 
IBRNet$_{ft-1.0h}$~\cite{ibrnet} & 23.92 & 0.816 & 0.229 & 20.51 & 0.810 & 0.212 \\
MVSNeRF$_{ft-15min}$~\cite{mvsnerf} & 21.34 & 0.831 & 0.253 & 22.32 & 0.850 & 0.217 \\
ENeRF$_{ft-1.0h}$~\cite{enerf} & 24.35 & 0.884 & \underline{0.148} & \underline{24.01} & \underline{0.885} & \underline{0.141} \\
3D-GS$_{ft-2min30s}$~\cite{3Dgaussians} & 21.07 & 0.825 & 0.255 & 19.19 & 0.731 & 0.384 \\
3D-GS$_{ft-15min}$~\cite{3Dgaussians} & \textbf{24.89} & \underline{0.897} & 0.152 & 22.42 & 0.838 & 0.215 \\
Ours$_{ft-90s}$ & \underline{24.73} & \textbf{0.910} & \textbf{0.133} & \textbf{24.43} & \textbf{0.896} & \textbf{0.140} \\
\midrule

\end{tabular}}
\end{table}

\begin{table}
\caption{\textbf{Quantitative per-scene breakdown results on the Real Forward-facing dataset.} PixelSplat$^*$ and Ours$^*$ represent the results obtained with a 2-view input and low-resolution images, while the other generalizable results are obtained with a 3-view input.}
\label{tab:llff_break}
\centering
\resizebox{0.75\linewidth}{!}{
\begin{tabular}{@{}l|cccccccc@{}}
\toprule
Scene & Fern & Flower & Fortress & Horns & Leaves & Orchids & Room & Trex \\
\midrule
Metric & \multicolumn{8}{c}{PSNR $\uparrow$} \\
\midrule
PixelNeRF~\cite{pixelnerf} & 12.40 & 10.00 & 14.07 & 11.07 & 9.85 & 9.62 & 11.75 & 10.55 \\
IBRNet~\cite{ibrnet} & 20.83 & 22.38 & 27.67 & 22.06 & 18.75 & 15.29 & 27.26 & 20.06 \\
MVSNeRF~\cite{mvsnerf} & 21.15 & \underline{24.74} & 26.03 & 23.57 & 17.51 & 17.85 & 26.95 & \textbf{23.20} \\
ENeRF~\cite{enerf} & \underline{21.92} & 24.28 & \underline{30.43} & \underline{24.49} & \underline{19.01} & \underline{17.94} & \underline{29.75} & 21.21 \\
MatchNeRF~\cite{matchnerf} & 20.98 & 23.97 & 27.44 & 23.14 & 18.62 & \textbf{18.07} & 26.77 & 20.47 \\
Ours & \textbf{22.45} & \textbf{25.66} & \textbf{30.46} & \textbf{24.70} & \textbf{19.81} & 17.86 & \textbf{29.86} & \underline{21.75} \\
\midrule
PixelSplat$^{*}$~\cite{pixelsplat} & 22.41 & 24.48 & 27.00 & 25.02 & 19.80 & 18.39 & 27.56 & 19.28 \\
Ours$^{*}$ & 22.47 & 23.96 & 30.00 & 23.97 & 19.42 & 17.06 & 28.59 & 20.95 \\
\midrule
NeRF$_{ft-10.2h}$~\cite{nerf} & 23.87 & 26.84 & \textbf{31.37} & 25.96 & 21.21 & 19.81 & \textbf{33.54} & \textbf{25.19}  \\
IBRNet$_{ft-1.0h}$~\cite{ibrnet} & 22.64 & 26.55 & 30.34 & 25.01 & \underline{22.07} & 19.01 & 31.05 & 22.34 \\ 
MVSNeRF$_{ft-15min}$~\cite{mvsnerf} & 23.10 & 27.23 & 30.43 & \underline{26.35} & 21.54 & 20.51 & 30.12 & 24.32 \\
ENeRF$_{ft-1.0h}$~\cite{enerf} & 22.08 & \textbf{27.74} & 29.58 & 25.50 & 21.26 & 19.50 & 30.07 & 23.39 \\
3D-GS$_{ft-2min}$~\cite{3Dgaussians} & \textbf{24.62} & 23.23 & 28.94 & 20.49 & 15.81 & \textbf{22.76} & 22.17 & 19.19 \\
3D-GS$_{ft-10min}$~\cite{3Dgaussians} & \underline{24.58} & 24.90 & 29.27 & 21.90 & 15.77 & \underline{22.42} & 31.45 & 21.09 \\
Ours$_{ft-45s}$ & 24.32 & \underline{27.66} & \underline{31.05} & \textbf{30.30} & \textbf{22.53} & 22.38 & \underline{33.11} & \underline{24.51} \\
\midrule
Metric & \multicolumn{8}{c}{SSIM $\uparrow$} \\
\midrule
PixelNeRF~\cite{pixelnerf} & 0.531 & 0.433 & 0.674 & 0.516 & 0.268 & 0.317 & 0.691 & 0.458 \\
IBRNet~\cite{ibrnet} & 0.710 & 0.854 & \underline{0.894} & 0.840 & 0.705 & 0.571 & 0.950 & 0.768 \\
MVSNeRF~\cite{mvsnerf} & 0.638 & 0.888 & 0.872 & 0.868 & 0.667 & 0.657 & 0.951 & \textbf{0.868}  \\
ENeRF~\cite{enerf} & \underline{0.774} & \underline{0.893} & \textbf{0.948} & \underline{0.905} & \underline{0.744} & \underline{0.681} & \underline{0.971} & 0.826 \\
MatchNeRF~\cite{matchnerf} & 0.726 & 0.861 & 0.906 & 0.870 & 0.690 & 0.675 & 0.949 & 0.767 \\
Ours & \textbf{0.792} & \textbf{0.908} & \textbf{0.948} & \textbf{0.913} & \textbf{0.784} & \textbf{0.701} & \textbf{0.973} & \underline{0.841} \\
\midrule
PixelSplat$^{*}$~\cite{pixelsplat} & 0.754 & 0.868 & 0.891 & 0.884 & 0.747 & 0.673 & 0.952 & 0.712 \\
Ours$^{*}$ & 0.787 & 0.877 & 0.937 & 0.896 & 0.772 & 0.649 & 0.962 & 0.798 \\
\midrule
NeRF$_{ft-10.2h}$~\cite{nerf} & 0.828 & 0.897 & \underline{0.945} & 0.900 & 0.792 & 0.721 & \underline{0.978} & \underline{0.899} \\
IBRNet$_{ft-1.0h}$~\cite{ibrnet} & 0.774 & 0.909 & 0.937 & 0.904 & \underline{0.843} & 0.705 & 0.972 & 0.842 \\ 
MVSNeRF$_{ft-15min}$~\cite{mvsnerf} & 0.795 & 0.912 & 0.943 & \underline{0.917} & 0.826 & 0.732 & 0.966 & 0.895 \\
ENeRF$_{ft-1.0h}$~\cite{enerf} & 0.770 & \underline{0.923} & 0.940 & 0.904 & 0.827 & 0.725 & 0.965 & 0.869 \\
3D-GS$_{ft-2min}$~\cite{3Dgaussians} & \textbf{0.845} & 0.850 & 0.918 & 0.813 & 0.495 & \textbf{0.850} & 0.930 & 0.759 \\
3D-GS$_{ft-10min}$~\cite{3Dgaussians} & 0.841 & 0.870 & 0.934 & 0.820 & 0.490 & 0.843 & 0.975 & 0.807 \\
Ours$_{ft-45s}$ & \underline{0.835} & \textbf{0.937} & \textbf{0.963} & \textbf{0.962} & \textbf{0.871} & \underline{0.844} & \textbf{0.986} & \textbf{0.911} \\
\midrule
Metric & \multicolumn{8}{c}{LPIPS $\downarrow$} \\
\midrule
PixelNeRF~\cite{pixelnerf} & 0.650 & 0.708 & 0.608 & 0.705 & 0.695 & 0.721 & 0.611 & 0.667 \\
IBRNet~\cite{ibrnet} & 0.349 & 0.224 & 0.196 & 0.285 & 0.292 & 0.413 & 0.161 & 0.314 \\
MVSNeRF~\cite{mvsnerf} & 0.238 & 0.196 & 0.208 & 0.237 & 0.313 & \textbf{0.274} & 0.172 & \underline{0.184} \\
ENeRF~\cite{enerf} & \underline{0.224} & \underline{0.164} & \textbf{0.092} & \underline{0.161} & \underline{0.216} & 0.289 & \underline{0.120} & 0.192 \\
MatchNeRF~\cite{matchnerf} & 0.285 & 0.202 & 0.169 & 0.234 & 0.277 & 0.325 & 0.167 & 0.294 \\
Ours & \textbf{0.193} & \textbf{0.133} & \underline{0.096} & \textbf{0.148} & \textbf{0.189} & \underline{0.275} & \textbf{0.104} & \textbf{0.177} \\
\midrule
PixelSplat$^{*}$~\cite{pixelsplat} & 0.181 & 0.158 & 0.149 & 0.160 & 0.214 & 0.275 & 0.128 & 0.258 \\
Ours$^{*}$ & 0.173 & 0.124 & 0.082 & 0.142 & 0.182 & 0.261 & 0.083 & 0.167 \\
\midrule
NeRF$_{ft-10.2h}$~\cite{nerf} & 0.291 & 0.176 & 0.147 & 0.247 & 0.301 & 0.321 & 0.157 & 0.245 \\
IBRNet$_{ft-1.0h}$~\cite{ibrnet} & 0.266 & 0.146 & 0.133 & 0.190 & 0.180 & 0.286 & \underline{0.089} & 0.222 \\ 
MVSNeRF$_{ft-15min}$~\cite{mvsnerf} & 0.253 & 0.143 & 0.134 & 0.188 & 0.222 & 0.258 & 0.149 & 0.187 \\
ENeRF$_{ft-1.0h}$~\cite{enerf} & 0.197 & \underline{0.121} & \underline{0.101} & \underline{0.155} & \underline{0.168} & 0.247 & 0.113 & \underline{0.169}  \\
3D-GS$_{ft-2min}$~\cite{3Dgaussians} & \underline{0.154} & 0.204 & 0.146 & 0.338 & 0.425 & \textbf{0.142} & 0.222 & 0.309 \\
3D-GS$_{ft-10min}$~\cite{3Dgaussians} & \textbf{0.147} & 0.183 & 0.121 & 0.289 & 0.421 & \underline{0.145} & 0.123 & 0.276\\
Ours$_{ft-45s}$ & 0.161 & \textbf{0.097} & \textbf{0.077} & \textbf{0.091} & \textbf{0.143} & \underline{0.145} & \textbf{0.079} & \textbf{0.113} \\
\bottomrule
\end{tabular}}
\end{table}

\begin{table}
\caption{\textbf{Quantitative per-scene breakdown results on the NeRF Synthetic dataset.} PixelSplat$^*$ and Ours$^*$ represent the results obtained with a 2-view and low-resolution input, while the other generalizable results are obtained with a 3-view input.}
\label{tab:nerf_break}
\centering
\resizebox{0.75\linewidth}{!}{
\begin{tabular}{@{}l|cccccccc@{}}
\toprule
Scene & Chair & Drums & Ficus & Hotdog & Lego & Materials & Mic & Ship\\
\midrule
Metric & \multicolumn{8}{c}{PSNR $\uparrow$} \\
\midrule
PixelNeRF~\cite{pixelnerf} & 7.18 & 8.15 & 6.61 & 6.80 & 7.74 & 7.61 & 7.71 & 7.30  \\
IBRNet~\cite{ibrnet} & 24.20 & 18.63 & 21.59 & 27.70 & 22.01 & 20.91 & 22.10 & 22.36 \\
MVSNeRF~\cite{mvsnerf} & 23.35 & 20.71 & 21.98 & 28.44 & 23.18 & 20.05 & 22.62 & 23.35 \\
ENeRF~\cite{enerf} & \underline{28.29} & \underline{21.71} & \textbf{23.83} & \underline{34.20} & \textbf{24.97} & \underline{24.01} & \underline{26.62} & \underline{25.73} \\
MatchNeRF~\cite{matchnerf} & 25.23 & 19.97 & 22.72 & 24.19 & \underline{23.77} & 23.12 & 24.46 & 22.11 \\
Ours & \textbf{28.93} & \textbf{22.20} & \underline{23.55} & \textbf{35.01} & \textbf{24.97} & \textbf{24.49} & \textbf{26.80} & \textbf{25.75} \\
\midrule
PixelSplat$^{*}$~\cite{pixelsplat} & 16.45 & 15.40 & 17.47 & 13.25 & 16.86 & 15.88 & 16.83 & 14.06 \\
Ours$^{*}$ & 27.95 & 21.20 & 23.22 & 33.79 & 24.23 & 24.55 & 24.22 & 23.54 \\
\midrule
NeRF~\cite{nerf} & 31.07 & 25.46 & 29.73 & 34.63 & 32.66 & \textbf{30.22} & 31.81 & 29.49  \\
IBRNet$_{ft-1.0h}$~\cite{ibrnet} & 28.18 & 21.93 & 25.01 & 31.48 & 25.34 & 24.27 & 27.29 & 21.48 \\ 
MVSNeRF$_{ft-15min}$~\cite{mvsnerf} & 26.80 & 22.48 & 26.24 & 32.65 & 26.62 & 25.28 & 29.78 & 26.73 \\
ENeRF$_{ft-1.0h}$~\cite{enerf} & 28.94 & 25.33 & 24.71 & \underline{35.63} & 25.39 & 24.98 & 29.25 & 26.36  \\
3D-GS$_{ft-1min15s}$~\cite{3Dgaussians} & \underline{31.90} & \textbf{26.56} & \textbf{34.21} & 34.21 & \textbf{36.28} & 29.80 & \textbf{34.56} & \underline{29.70} \\
3D-GS$_{ft-7min}$~\cite{3Dgaussians} & 31.20 & \underline{26.26} & \underline{33.93} & 34.30 & \underline{36.10} & 29.53 & \underline{34.39} & 28.90 \\
Ours$_{ft-50s}$ & \textbf{32.80} & 25.91 & 31.54 & \textbf{36.85} & 35.68 & \underline{29.83} & 33.92 & \textbf{31.09} \\
\midrule
Metric & \multicolumn{8}{c}{SSIM $\uparrow$} \\
\midrule
PixelNeRF~\cite{pixelnerf} & 0.624 & 0.670 & 0.669 & 0.669 & 0.671 & 0.644 & 0.729 & 0.584 \\
IBRNet~\cite{ibrnet} & 0.888 & 0.836 & 0.881 & 0.923 & 0.874 & 0.872 & 0.927 & 0.794 \\
MVSNeRF~\cite{mvsnerf} & 0.876 & 0.886 & 0.898 & 0.962 & 0.902 & 0.893 & 0.923 & 0.886 \\
ENeRF~\cite{enerf} & \underline{0.965} & \underline{0.918} & \underline{0.932} & \underline{0.981} & \underline{0.948} & \underline{0.937} & \underline{0.969} & \underline{0.891} \\
MatchNeRF~\cite{matchnerf} & 0.908 & 0.868 & 0.897 & 0.943 & 0.903 & 0.908 & 0.947 & 0.806 \\
Ours & \textbf{0.969} & \textbf{0.927} & \textbf{0.935} & \textbf{0.984} & \textbf{0.953} & \textbf{0.946} & \textbf{0.974} & \textbf{0.895} \\
\midrule
PixelSplat$^{*}$~\cite{pixelsplat} & 0.816 & 0.787 & 0.857 & 0.644 & 0.799 & 0.764 & 0.861 & 0.508 \\
Ours$^{*}$ & 0.962 & 0.909 & 0.920 & 0.978 & 0.940 & 0.940 & 0.957 & 0.873 \\
\midrule
NeRF~\cite{nerf} & 0.971 & 0.943 & 0.969 & 0.980 & 0.975 & \underline{0.968} & 0.981 & 0.908 \\
IBRNet$_{ft-1.0h}$~\cite{ibrnet} & 0.955 & 0.913 & 0.940 & 0.978 & 0.940 & 0.937 & 0.974 & 0.877 \\ 
MVSNeRF$_{ft-15min}$~\cite{mvsnerf} & 0.934 & 0.898 & 0.944 & 0.971 & 0.924 & 0.927 & 0.970 & 0.879 \\
ENeRF$_{ft-1.0h}$~\cite{enerf} & 0.971 & \textbf{0.960} & 0.939 & \underline{0.985} & 0.949 & 0.947 & 0.985 & 0.893  \\
3D-GS$_{ft-1min15s}$~\cite{3Dgaussians} & \underline{0.981} & \underline{0.956} & \textbf{0.986} & 0.983 & \underline{0.987} & \textbf{0.970} & \underline{0.991} & \underline{0.918} \\
3D-GS$_{ft-7min}$~\cite{3Dgaussians} & 0.977 & 0.951 & \underline{0.985} & 0.981 & \underline{0.987} & \underline{0.968} & \textbf{0.992} & 0.909 \\
Ours$_{ft-50s}$ & \textbf{0.983} & 0.952 & 0.981 & \textbf{0.987} & \textbf{0.988} & \textbf{0.970} & \textbf{0.992} & \textbf{0.921} \\
\midrule
Metric & \multicolumn{8}{c}{LPIPS $\downarrow$} \\
\midrule
PixelNeRF~\cite{pixelnerf} & 0.386 & 0.421 & 0.335 & 0.433 & 0.427 & 0.432 & 0.329 & 0.526 \\
IBRNet~\cite{ibrnet} & 0.144 & 0.241 & 0.159 & 0.175 & 0.202 & 0.164 & 0.103 & 0.369\\
MVSNeRF~\cite{mvsnerf} & 0.282 & 0.187 & 0.211 & 0.173 & 0.204 & 0.216 & 0.177 & 0.244 \\
ENeRF~\cite{enerf} & \underline{0.055} & \underline{0.110} & \underline{0.076} & \underline{0.059} & \underline{0.075} & \underline{0.084} & \underline{0.039} & \underline{0.183} \\
MatchNeRF~\cite{matchnerf} & 0.107 & 0.185 & 0.117 & 0.162 & 0.160 & 0.119 & 0.060 & 0.398 \\
Ours & \textbf{0.036} & \textbf{0.091} & \textbf{0.069} & \textbf{0.040} & \textbf{0.066} & \textbf{0.063} & \textbf{0.027} & \textbf{0.179} \\
\midrule
PixelSplat$^{*}$~\cite{pixelsplat} & 0.260 & 0.287 & 0.282 & 0.365 & 0.273 & 0.309 & 0.241 & 0.493 \\
Ours$^{*}$ & 0.039 & 0.098 & 0.066 & 0.038 & 0.071 & 0.050 & 0.038 & 0.170 \\
\midrule
NeRF~\cite{nerf} & 0.055 & 0.101 & 0.047 & 0.089 & \underline{0.054} & 0.105 & 0.033 & 0.263 \\
IBRNet$_{ft-1.0h}$~\cite{ibrnet} & 0.079 & 0.133 & 0.082 & 0.093 & 0.105 & 0.093 & 0.040 & 0.257 \\ 
MVSNeRF$_{ft-15min}$~\cite{mvsnerf} & 0.129 & 0.197 & 0.171 & 0.094 & 0.176 & 0.167 & 0.117 & 0.294 \\
ENeRF$_{ft-1.0h}$~\cite{enerf} & 0.030 & \textbf{0.045} & 0.071 & \textbf{0.028} & 0.070 & 0.059 & 0.017 & 0.183\\
3D-GS$_{ft-1min15s}$~\cite{3Dgaussians} & \underline{0.022} & \underline{0.059} & \textbf{0.016} & 0.042 & \textbf{0.021} & \underline{0.041} & \underline{0.010} & 0.180 \\
3D-GS$_{ft-7min}$~\cite{3Dgaussians} & 0.026 & 0.062 & \underline{0.018} & 0.044 & \textbf{0.021} & 0.043 & \textbf{0.009} & \underline{0.172} \\
Ours$_{ft-50s}$ & \textbf{0.021} & \underline{0.059} & 0.022 & \underline{0.032} & \textbf{0.021} & \textbf{0.038} & \underline{0.010} & \textbf{0.138} \\
\bottomrule
\end{tabular}}
\end{table}
\end{document}